\pdfoutput=1

\documentclass[11pt]{article}

\usepackage{acl}

\usepackage{times}
\usepackage{latexsym}

\usepackage{graphicx}
\usepackage{algorithm}
\usepackage{algpseudocode}
\usepackage{xcolor}
\usepackage{multirow}
\usepackage{etoolbox}
\usepackage{balance}
\usepackage{amsmath}
\usepackage{bbm}
\usepackage{amsfonts}
\usepackage{comment}

\usepackage{pifont}
\usepackage{stfloats}
\usepackage{subfigure}

\usepackage{booktabs} %

\usepackage[T1]{fontenc}

\usepackage[utf8]{inputenc}

\usepackage{microtype}

\usepackage{inconsolata}

\usepackage{xspace}
\newcommand{\ours}{STUN\xspace}

\usepackage{etoolbox} %

\newtoggle{showprev}
\toggletrue{showprev}

\newtoggle{shownew}
\toggletrue{shownew}

\usepackage{array} %
\makeatletter
\newcommand*{\@rowstyle}{}
\newcommand*{\rowstyle}[1]{%
  \gdef\@rowstyle{#1}%
  \@rowstyle\ignorespaces%
}
\newcolumntype{=}{%
  >{\gdef\@rowstyle{}}%
}
\newcolumntype{+}{%
  >{\@rowstyle}%
}
\makeatother

\definecolor{Emerald}{RGB}{192, 41, 66}
\iftoggle{shownew}{
    
    \newcommand{\jsleeobj}[1]{ %
    {\let\Cap\caption
    \def\caption##1{\Cap{\color{Emerald}##1}}
    \color{Emerald}#1}
    }
}
{
    
    \newcommand{\jsleeobj}[1]{#1}
    
}

\definecolor{Gray}{RGB}{200, 200, 200}
\iftoggle{showprev}{
    \newcommand{\jsleeprev}[1]{{\color{Gray}#1}}
    \newcommand{\jsleeprevobj}[1]{ %
    {\let\Cap\caption
    \def\caption##1{\Cap{\color{Gray}##1}}
    \color{Gray}#1}
    }
    \newcommand{\jsleeprevrow}[1]{\rowstyle{\color{Gray}}#1}
}
{
    \newcommand{\jsleeprev}[1]{}
    \newcommand{\jsleeprevobj}[1]{}
    \newcommand{\jsleeprevrow}[1]{}
}

\title{
  \includegraphics[height=1em]{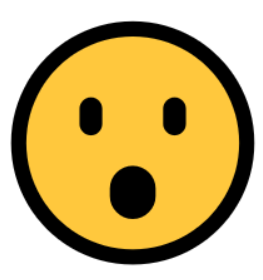}
STUN: Structured-Then-Unstructured Pruning \\
for Scalable MoE Pruning
}

 \author{\textbf{
Jaeseong Lee\textsuperscript{*},
Seung-won Hwang\thanks{Work done while visiting Snowflake. Correspond to seungwonh@snu.ac.kr},
Aurick Qiao,}\\
\textbf{Daniel Campos,
Zhewei Yao,
Yuxiong He}
\\
Snowflake AI Research, Seoul National University\textsuperscript{\rm *}\\
 }

\begin{document}
\maketitle
\begin{abstract}
Mixture-of-experts (MoEs) 
have been adopted to reduce inference costs by sparsely activating experts
in large language models (LLMs).
Despite these reductions, the massive number of parameters
in MoEs still makes them expensive to serve.
Conventionally, unstructured or structured pruning has been considered to reduce the number of parameters.
Our key contribution is exploring the interpolation between structured and unstructured pruning, to propose a novel structured-then-unstructured (STUN) approach outperforming both structured and unstructured pruning, especially for MoEs.
In the first stage, we show a scalable  
 expert pruning with O(1) forward pass, unlike existing work requiring  $O(\frac{k^n}{\sqrt{n}})$ forward passes for $n$ experts that
cannot scale for recent MoEs with hundreds of experts.
We then show our expert-pruned MoEs are robust to unstructured pruning to follow. 
Experiments on Snowflake Arctic and Mixtral show that our proposal is highly effective-- For Snowflake Arctic, a 480B-sized MoE with 128 experts, our method needs only one H100 and two hours to achieve nearly no loss in performance with 40\% sparsity, even in generative tasks such as GSM8K, where state-of-the-art structured or unstructured pruning methods fail. The code is publicly available.\footnote{https://github.com/thnkinbtfly/STUN}

\end{abstract}

\section{Introduction}
Large language models (LLMs) have become state-of-the-art for various tasks~\cite{GPT42023openai,LLaMA2Llama2023touvron,Mistral2023jiang,Jamba2024lieber}. 
However, their prohibitive inference cost is becoming a bottleneck to deployment~\cite{Challenges2023kaddour}, and detrimental to the environment~\cite{Energy2019strubell,GreenPLM2023zeng}.

Mixture-of-experts (MoE) presents a promising alternative, %
by sparsely activating a specific subset of parameters, named as experts, to reduce the inference cost. This architecture has been empirically proven effective, in 
training cost~\cite{Switch2022fedus}, and inference cost~\cite{GLaM2022du}.

Despite these reductions, the massive number of parameters remains unchanged, requiring significantly more GPU memory,
which makes serving large MoE models challenging for many.
Additionally, recent MoEs tend to increase the number of experts $n$, resulting in even larger MoEs.
For instance, accommodating 56B parameters of Mixtral~\cite{Mixtral2024jiang} with 8 experts or 132B of DBRX~\cite{databricks2024databricks} with 16 experts, or 480B of Snowflake Arctic~\cite{SnowflakeLabs2024snowflake} with 128 experts, requires an ever-growing amount of memory and more GPUs to serve.

\begin{figure}[t]
    \centering
   \includegraphics[width=\columnwidth]{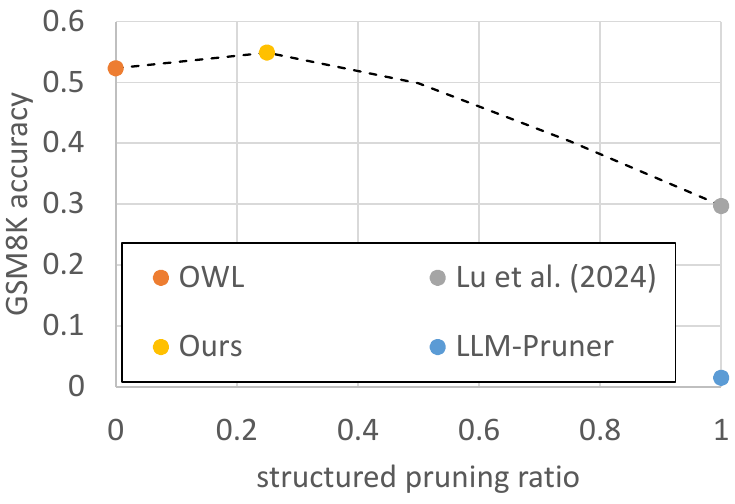}
   \caption{GSM8K 5-shot accuracy by pruning Mixtral-8x7B-Instruct by 50\% of sparsity. We probe interpolation of structured and unstructured pruning, by varying the ratio of structured pruning.}
    \label{fig:pruning_ratio_comparison}
\end{figure}

To reduce the number of parameters, unstructured~\cite{SparseGPT2023frantar,Wandasimple2024sun}, or structured pruning~\cite{LLMPruner2023ma} can be considered.
Unstructured pruning allows weight tensors to be sparse anywhere, while structured pruning
imposes patterns on sparsification, such as removing rows, entire weight tensors~\cite{LLMPruner2023ma}, or pruning experts in MoE~\cite{Not2024lu}.

In this paper, we propose the interpolation of two, Structured-Then-UNstructured pruning (\ours).
\autoref{fig:pruning_ratio_comparison} motivates our interpolated method, where unstructured- or structured-only, $x=0$ and $x=1$, respectively, is outperformed by the peak, combining both.

Our first phase, 
expert  pruning, by leveraging model-inherent expert for pruning, significantly outperforms
row/column-level structured pruning (\autoref{fig:pruning_ratio_comparison} blue; \citealp{LLMPruner2023ma})
However, existing expert-level pruning for MoE (\autoref{fig:pruning_ratio_comparison} grey; \citealp{Not2024lu}) often does not scale well over the solution space,
requiring an exhaustive combination of experts, leading to $O(\frac{k^n}{\sqrt{n}})$ GPU calls, with $k=\frac{1}{\phi^\phi (1-\phi)^{1-\phi}}$, and $\phi<1$ is sparsity~\cite{Not2024lu}. While this was
acceptable in an early MoE work with
few experts,
it does not scale to recent trends in MoEs with large $n$~\cite{Qwen2023bai,DeepSeekMoE2024dai,SnowflakeLabs2024snowflake}, or even  infinity~\cite{Mixture2024he}. 
Our distinction is drastically reducing the number of GPU calls to $O(1)$, without compromising the performance. The main intuition is leveraging a latent structure between experts, based on behavior similarity, such that the greedy decision of whether to prune closely captures the joint
pruning effect.

The second contribution is 
allowing unstructured phase to follow, to consider both inter- and intra-expert sparsity (\autoref{tab:taxonomy}). \ours removes redundant experts by expert-level structured pruning first, then desires fine-grained sparsity within individual experts.\footnote{From now on, we will define sparsity as the number of pruned parameters divided by the total number of parameters in the original model.}

We support \ours with the findings of \citet{What2024mason-williams}, which show that higher kurtosis in the weight distribution (indicating many outliers) suggests more weights can be pruned while maintaining performance, highlighting the robustness of unstructured pruning. We argue that expert-level pruning does not reduce kurtosis, thereby preserving the network's resilience to unstructured pruning.

\definecolor{mygrey}{RGB}{165,165,165}
\definecolor{mysky}{RGB}{91,155,213}
\definecolor{myyellow}{RGB}{255,192,0}
\definecolor{myorange}{RGB}{237,125,49}
\begin{table*}[]
\centering
\begin{tabular}{cc|cc}
\hline
Inter-Expert Sparsity & Intra-Expert Sparsity &  \autoref{fig:pruning_ratio_comparison}  \\ \hline
  \ding{51}           &   \ding{55}    &                            \citealp{Not2024lu} ({\color{mygrey}$\bullet $})     \\
  \ding{55}           &   \ding{51}    &                            LLM-Pruner ({\color{mysky}$\bullet $}), Wanda, OWL ({\color{myorange}$\bullet $})      \\
   \ding{51}           &   \ding{51}    &                          Ours ({\color{myyellow}$\bullet $})      \\ \hline
\end{tabular}
\caption{Comparison of existing pruning methods for MoEs.}
\label{tab:taxonomy}
\end{table*}

Our contributions can be summarized as follows:

\begin{itemize}
    \item  We propose \ours, the first method to combine structured and unstructured pruning, outperforming both approaches.

    \item \textbf{Scalable first phase:} We design an expert-level pruning method with $O(1)$ GPU calls, outperforming the $O(\frac{k^n}{\sqrt{n}})$ solution~\cite{Not2024lu}.
        \item \textbf{Justifying the second phase:} We show the expert-pruned network remains robust to unstructured pruning to follow.
\item \textbf{State-of-the-art efficiency and compression:} For Snowflake Arctic (480B, 128 experts), it requires just 1 H100 and two hours, with no backpropagation or fine-tuning needed.
Compression reaches up to
 40\% sparsity without compromising performance, even in generative tasks like GSM8K, where unstructured pruning fails. We report consistent results for Mixtral models.
\end{itemize}

\section{Related Work}

\subsection{LLM Pruning}

LLM pruning can be classified into unstructured and structured pruning~\cite{Pruning2021behnke}. Unstructured pruning involves finding mask tensors to sparsify weight tensors. 
Such masking leads to practical speedups in hardware such as CPU~\cite{Deepsparseneuralmagic2021neuralmagic}, and ongoing research is actively developing methods to achieve similar speedups on GPUs~\cite{Accelerating2021mishra,exploring2024zhao}.
SparseGPT~\cite{SparseGPT2023frantar} uses the Hessian matrix for second-order Taylor approximation, while GBLM-Pruner~\cite{GBLMSize2024das} and Pruner-Zero~\cite{PrunerZero2024dong} leverage gradients to identify mask tensors. However, as these methods demand substantial GPU memory for LLMs,
we focus on  more memory-efficient approaches, using two recent baselines:
Wanda~\cite{Wandasimple2024sun} evaluates the \textit{importance} of neurons in each layer by its weight multiplied by the activation value, removing those with low scores. While Wanda assumes a uniform sparsity across layers, OWL~\cite{OWLOutlier2024yin} probes the optimal sparsity per layer, given the pruning budget.

Structured pruning, on the other hand, imposes constraints on the sparsification pattern, such as  removing rows, columns, or even entire weight tensors. Early methods that involve pruning attention heads~\cite{Analyzing2019voita,Layerwise2021shim,Know2021zhang}, rows~\cite{Adaptive2022gong}, entire dense layers~\cite{Super2021liang}, or whole transformer blocks~\cite{Reducing2019fan,Deep2020li} fall under this category. 
Recent works have applied structured pruning for LLMs~\cite{LLMPruner2023ma,MINILLM2024cheng,Optimizationbased2024gao,Structured2024zhang,Everybody2024dery}, but without fine-tuning,  these methods generally underperform when compared to unstructured pruning.

Our distinction is to introduce a new class of pruning-- structured-then-unstructured pruning-- and demonstrate its significant advantages for MoEs, 
surpassing the performance of either method alone. This approach differs from previous methods that combine structured and unstructured pruning~\cite{oBERTOptimal2022kurtic}, which failed to outperform unstructured pruning.

\subsection{Expert Pruning}
Early work on expert pruning was domain-specific~\cite{Scalable2021kim,Memoryefficient2023koishekenov,EEPEfficient2024liu}, such as in translation MoEs, by keeping most activated experts~\cite{Scalable2021kim}, or pruning based on gate statistics~\cite{Memoryefficient2023koishekenov}.
\citet{Not2024lu} introduced a domain-agnostic expert pruning, 
searching for the best combination of experts to reduce the reconstruction loss, 
and quantify their criticality in output prediction. Concurrently to our study, \cite{Diversifying2024zhang} proposed an efficient expert pruning method.

Our distinction is two-fold. First, we interpolate expert pruning with unstructured pruning to outperform either method alone.
Second, for scalable expert-level structured pruning, 
we derive a scalable expert-level structured pruning method with $O(1)$ GPU calls, 
improving on the $O(\frac{k^n}{\sqrt{n}})$ solution enumerating combinatorial pruning.

\subsection{Pruning Robustness}
Robustness in post-hoc pruning  is quantified by whether performance is maintained after pruning. 
Kurtosis of weights~\cite{What2024mason-williams} has been used as a proxy for robustness, with networks showing higher weight kurtosis able to tolerate higher unstructured pruning ratios. Our contribution is demonstrating that an expert-pruned network remains robust to additional unstructured pruning, which naturally supports our design of unstructured pruning as the second phase.
\section{Preliminaries: MoE}

As a promising alternative to large language models, which incur prohibitive inference costs, MoE employs a multitude of specialized experts. In each forward pass, MoE selectively activates specific experts conditioned on input tokens, thereby reducing the train and inference costs.

We now formally describe the behavior of an MoE. %
An MoE layer $M$ consists of experts $E_i=E(x;\theta_i)$, where $\theta_i$ represents the parameters of expert $E_i$, and a router layer $r$. Each expert $E$ typically follows the same MLP architecture.

First, the router layer selects which experts to sparsely activate based on the current input token, and provides the coefficients $r(x) \in \mathbb{R}^n$ for linear combination of selected expert outputs.
The coefficients $r(x) $ and the top-k indices of experts $\mathcal{T}$ are formulated as follows:
\begin{gather}
r(x) = \mathrm{softmax}(W x)\\
\mathcal{T} = \mathrm{topk}(r(x))
\end{gather}
where $W$ is the learnable weight matrix for router $r$.

Next, these coefficients are used for the linear combination of expert outputs:
\begin{equation}\label{eq:MoE}
M(x;\mathbf{\theta}) = \sum_{i \in \mathcal{T}} r_i (x) E (x;\theta_i)
\end{equation}

\section{Proposed Method}\label{sec:ours}

\begin{figure}[t]
    \centering
   \includegraphics[width=\columnwidth]{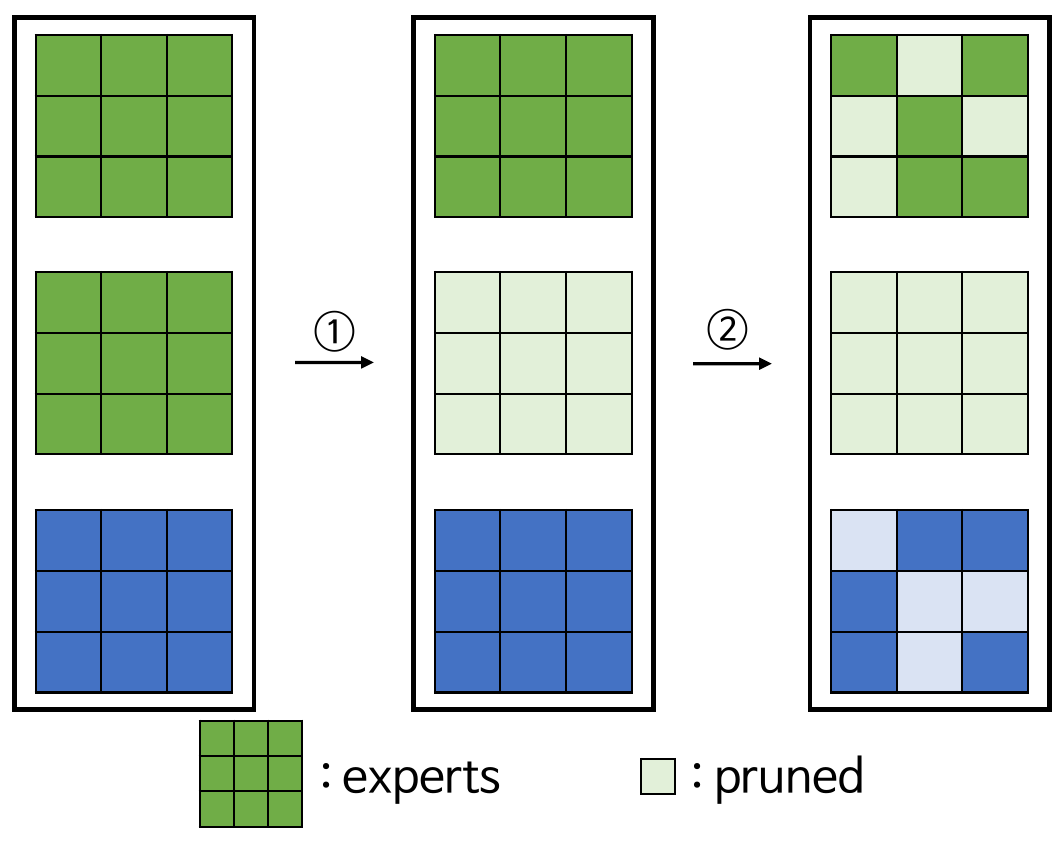}
   \caption{Overview of our proposed \ours. \ding{192} We first remove redundant experts with expert-level structured pruning, then \ding{193} perform unstructured pruning inside individual experts. Black box represents a layer in MoE, and different colors represent different behavioral similarities.}
    \label{fig:overview}
\end{figure}

\subsection{Overview: \ours}\label{sec:prop_overview}

\autoref{fig:overview} overviews our two-phase
approach, interpolating structured and unstructured pruning as motivated
in \autoref{tab:taxonomy}.
Section \ref{sec:o1_gpu_intro} describes 
\ding{192} how we remove redundant experts with expert-level structured pruning with high scalability, then Section \ref{sec:unstructured_kurtosis}
describes \ding{193} how we perform unstructured pruning inside individual experts.

\subsection{Expert-level Structured Pruning with $O(1)$ GPU calls}\label{sec:o1_gpu_intro}
Now, we describe our expert-level structured pruning with $O(1)$ GPU calls.\footnote{We focus on GPU cost, since it dominates the CPU cost-- For example, the accumulated CPU cost (including the hyperparameter search to meet desired sparsity) required by our algorithm is less than 1 minute, even on 480B Snowflake Arctic.}
Previous solution~\cite{Not2024lu} minimizes reconstruction loss (Section \ref{sec:combinatorial_reconstruction}), requiring GPU call per combination of experts, that is $O(\frac{k^n}{\sqrt{n}})$.
Our key contribution is to approximate this combinatorial reconstruction loss to reduce the number of GPU calls to $O(1)$,
by leveraging \textit{latent cluster structure} among experts, based on behavioral similarity.

Specifically, we find clusters of similar experts layer by layer, yielding a total of $\phi n l$ clusters in the whole MoE, where $\phi$ is the sparsity, $n$ is the number of experts in each layer, and $l$ is the number of layers in MoE.
Then we greedily prune every expert but one representative per each cluster.

Later sections show why our greedy pruning is as effective as its combinatorial counterpart. 

\subsubsection{$O(\frac{k^n}{\sqrt{n}})$: Combinatorial Reconstruction Loss}\label{sec:combinatorial_reconstruction}
We start from the conventional goal of pruning-- minimizing the reconstruction loss.
Reconstruction loss has been employed to assess how closely the pruned model  $\theta-\theta_S$
without expert set $S$
mirrors the behavior of
the unpruned $\theta$
~\cite{Not2024lu}.
Formally, this loss is quantified by
the Frobenius norm of the difference between the original output $M(x;\theta)$ and 
the output of pruned layer $M(x;\theta-\theta_S)$, denoted
as $\mathcal{E}_S$.
\begin{equation}\label{eq:reconstruction_loss}
    \mathcal{E}_S=\|M(x;\theta)-M(x;\theta-\theta_S)\|_F
\end{equation}
where $x$ is the whole input we consider.
The objective of pruning is to explore all possible combinations of experts,
 ${n}\choose{|S|}$, to determine the expert set $S$ that minimizes $\mathcal{E}_S$.

While such an exhaustive search is feasible for smaller models like
Mixtral~\cite{Mixtral2024jiang}, which contains only 8 experts, it becomes prohibitive for
recent MoEs featuring a massive number of experts.

To elaborate, deciding which experts to prune using Eq. 4 for $|S|=\phi n$ requires ${n\choose{|S|}} \approx O(\frac{k^n}{\sqrt{n}})$ forward passes according to Stirling's approximation, where $k=\frac{1}{\phi^\phi (1-\phi)^{1-\phi}}$, and $\phi<1$ represents the sparsity.
Our distinction is to lower
the computation to $O(1)$, without
compromising the performance-- In fact, as we will elaborate later, we outperform the combinatorial objective.

\subsubsection{Towards $O(n)$: Probabilistic Interpretation}
As a stepping stone towards $O(1)$, we propose to
rephrase the goal of finding $\theta_S$ to minimize $\mathcal{E}_S$ (Eq. \ref{eq:reconstruction_loss}) as:

\begin{multline}
\textrm{argmax}_S \prod_k P(X_k=s_k|X_1=s_1,\\
\cdots,X_{k-1}=s_{k-1})\label{eq:prob_conditional_2}
\end{multline}

Our contribution is greedy optimization without compromise for Eq. \ref{eq:prob_conditional_2}%
-- We decompose the multiplication of Eq.~\ref{eq:prob_conditional_2} at each step $k$, and obtain the distribution $P(X_k|s_1,\cdots,s_{k-1})$, to select $X_k$ that maximizes the probability.
To achieve it, we estimate the rank between the probabilities.
Such rank estimation can benefit from the latent structure among experts, specifically, a cluster of similar experts in MoE.
Given cluster mapping $c$ which maps an expert to a set of similarly behaving experts, we assign the value $P(E_i|S_{k-1})$, as follows:
\begin{equation}\label{eq:O_n_prob}
P(E_i|S_{k-1}) = \begin{cases}
    P(E_i) - p & c(E_i) \subseteq S_{k} \\
    P(E_i) & \mathrm{otherwise}
    \end{cases}
\end{equation}
This enables the calculation of all $P(E_i|S_{k-1})$ in Eq. \ref{eq:prob_conditional_2} from $P(E_i)$s, which needs only $n$ forwards in total.

\paragraph*{Clustering the Similar Experts} 
Our remaining task is to obtain
cluster information $c$: %
One signal is pairwise behavioral similarity $b_{i,j}$,
from the pretrained router weights $W$ at a minimal cost.
Suppose two rows $W_{i} \approx W_{j}$ are similar; then $r_i(x) \approx r_j(x)$, meaning $E_i,E_j$ tend to be selected by similar inputs, implying similar expertise.
Thus, %
the behavioral similarity $b_{i,j}$ between two experts $E_i,E_j$ can be obtained as follows:

\begin{equation}
b_{i,j} = - \|W_{i} - W_{j}\|_F\label{eq:sim_weight}
\end{equation}
which can be improved with coactivation statistics $a_{i,j}$, if we allow some inference cost.
As a result, we illustrate our clustering algorithm in Alg \ref{alg:cluster}, whose detailed derivation can be found in Appendix \ref{sec:O_n}.

\begin{algorithm}
\caption{Expert Clustering Algorithm}\label{alg:cluster}
\begin{algorithmic}
\Require $l \gets$ Number of layers
\Require $n \gets$ Number of experts per layer
\Require $\{a_{i,j}\}_{i,j} \gets$ Coactivation statistics of $E_i,E_j$ for every layer
\Require $\lambda_1, \lambda_2 \gets$ Hyperparameter for behavioral similarity
\Require $t \gets$ Threshold to determine sparsity
\Ensure $c \gets$ The mapping from expert to cluster of the similar experts
\For{$m$ in $[1..l]$}
\State $W \gets$ Router weight of layer $m$
\State $\{a_{i,j}\}_{i,j} \gets$ Coactivation statistics of layer $m$
\For{$i$ in $[1..n-1]$}
\For{$j$ in $[i+1..n]$}
\State $b_{i,j} \gets - \lambda_1 \|W_{i} - W_{j}\|_F + \lambda_2 a_{i,j}$
\EndFor
\EndFor
\For{$i$ in $[1..n]$}
\State $c(E_i) \gets \{E_i\}$
\EndFor
\While{$\min_{i,j} b_{i,j} < t$}
\State ${d,e} \gets \mathrm{argmin}_{i,j} b_{i,j}$
\State $m_d \gets \max_{i \in c(E_e)} b_{d,i}$
\State $m_e \gets \max_{i \in c(E_d)} b_{i,e}$
\If{$c(E_d) \neq c(E_e) \wedge \max(m_d,m_e)<t$}
\State $c(E_d)=c(E_e) \gets c(E_d)\cup c(E_e)$
\EndIf
\State $b_{d,e} \gets \infty$ \Comment{Mark as visited}
\EndWhile
\EndFor
\State \Return $c$
\end{algorithmic}
\end{algorithm}

\subsubsection{Towards $O(1)$: Taylor Approximation and Selective Reconstruction}\label{sec:O_1}
While the previous section immensely reduces the cost to obtain the probability distribution to $O(n)$ by requiring only $P(E_i)$s, we can 
further reduce the number of forward passes-- We aim to remove the GPU calls for $P(E_i)$, which is needed as in Eq. \ref{eq:joint_prob_construction}.

The key idea is approximating $E_i$'s reconstruction loss value $\mathcal{E}_i=\|M(x;\theta)-M(x;\theta-\theta_i)\|_F$.
To address this, with 1st order Taylor approximation, we find the expert closest to $\bar{\theta_i}$ within each cluster has the highest priority to be retained. We assign ranks similarly, and the same greedy algorithm is applied to optimize Eq. \ref{eq:prob_conditional_2}.
Additionally, we selectively reconstruct the expert.
The final algorithm is summarized in Alg \ref{alg:O1_prune}, which is described in detail in Appendix \ref{sec:O_1}.
\begin{algorithm}
\caption{Our $O(1)$ Expert Pruning}\label{alg:O1_prune}
\begin{algorithmic}
\Require $l \gets$ Number of layers
\Require $n \gets$ Number of experts per layer
\Require $c \gets$ The mapping from expert to cluster of the similar experts
\Require $\kappa \gets$ Threshold for selective reconstruction
\For{$m$ in $[1..l]$}
\State $r(m) = [~]$ 
\State $A \gets \{c(E_1),\cdots,c(E_n)\}$
\For{$C$ in $A$}
\State $\bar{\theta_i} \gets \frac{1}{|C|} \sum_{i\in C} \theta_i$

\If{$|A| < \kappa$}
\State $\theta_C \gets \bar{\theta_i}$ \Comment{Reconstruct}
\Else
\State $\theta_C \gets \min_{\theta_j \in C} \|\theta_j - \bar{\theta_i}\|_F$
\EndIf
\State $r(m).append(\theta_C)$
\EndFor
\EndFor
\State \Return $r$
\end{algorithmic}
\end{algorithm}

\subsection{Unstructured Pruning on Expert-pruned Model}\label{sec:unstructured_kurtosis}

Our main conjecture for STUN is
intra-expert sparsity yet remains intact after the first phase. Specifically, we propose to pursue fine-grained sparsity within the remaining experts,
by leveraging
unstructured pruning methods designed for general LLM , such as OWL~\cite{OWLOutlier2024yin} or Wanda~\cite{Wandasimple2024sun}. 

Now we theoretically verify our conjecture:  intra-expert sparsity remains high, or, even higher, after the expert pruning phase.
In other words, we explain why performing unstructured pruning after expert pruning is better than continuously performing unstructured pruning only.

Intra-expert sparsity, formally speaking, robustness to unstructured pruning, can be estimated by the kurtosis of weights~\cite{What2024mason-williams}.
Kurtosis is expressed as follows:

\begin{equation}
K(\theta) = E\left[\left(\frac{\theta-\mu}{\sigma}\right)^4\right]%
\end{equation}

Suppose the weight of experts $\theta$ follow a zero-meaned Gaussian distribution $\mathcal{N}$.
Unstructured pruning~\cite{Wandasimple2024sun,OWLOutlier2024yin,GBLMSize2024das,PrunerZero2024dong}, which tends to remove near-zero weights,\footnote{The importance score of unstructured pruning typically increases as the absolute value of the weight increases.} would shift the distribution closer to a bimodal symmetric distribution, whose kurtosis is minimum~\cite{Kurtosis1970darlington}. As a result, unstructured pruning would lower the kurtosis value, leaving less margin for further unstructured pruning.

In contrast, coarse structured pruning, such as expert pruning, is less likely to decrease the kurtosis value, since the assumption $\theta \sim \mathcal{N}$ still holds for remaining experts.
This implies that expert pruning preserves the robustness of unstructured pruning, unlike applying unstructured pruning with a similar sparsity.\footnote{In our experiments, the kurtosis increased from 14248 to 15623 after expert pruning.}

\section{Experiments}\label{sec:exp}

\begin{table*}[]
\centering
\setlength{\tabcolsep}{1.5pt}
\begin{tabular}{l|c|l|c|ccccc}
\hline
model                                                                               & sparsity              & method                          & GSM8K          & Avg(\textrightarrow)    & ARC-c & ARC-e & HellaSwag & MMLU  \\ \hline
\multirow{7}{*}{Arctic}                                                             & 0\%                   & unpruned                        & 70.74          & 68.33          & 56.91 & 84.60 & 66.94     & 64.86 \\ \cline{2-9} 
                                                                                    & \multirow{4}{*}{40\%} & \ours (w/ OWL)   & \textbf{70.28} & \textbf{67.66} & 57.68 & 83.29 & 64.94     & 64.75 \\
                                                                                    &                       & OWL                             & 63.76          & 67.35          & 56.74 & 84.13 & 65.08     & 63.47 \\ \cline{3-9} 
                                                                                    &                       & \ours (w/ Wanda) & \textbf{69.60} & \textbf{67.64} & 57.25 & 83.63 & 64.86     & 64.81 \\
                                                                                    &                       & Wanda                           & 64.59          & 67.54          & 57.00 & 84.64 & 65.19     & 63.32 \\ \cline{2-9} 
                                                                                           & \multirow{2}{*}{65\%} & \ours (w/ OWL)   & \textbf{43.97} & \textbf{62.67} & 51.54 & 80.01 & 59.91     & 59.24 \\
                                                                                    &                       & OWL                             & 13.42          & 56.68          & 44.37 & 76.64 & 53.69     & 52.02 \\ \hline
\multirow{3}{*}{\begin{tabular}[c]{@{}l@{}}Mixtral-8x7B\\ (Instruct)\end{tabular}}  & \multirow{3}{*}{65\%} & \ours (w/ OWL)   & \textbf{25.09} & \textbf{60.34} & 48.12 & 78.79 & 54.05     & 60.39 \\                                  &                       & OWL                             & 1.29           & 45.20          & 24.15 & 49.79 & 49.27     & 57.60 \\
                                                                                    &                       & LLM-Pruner                             & 1.29    & 31.74       & 22.78          & 45.96 & 35.09 & 23.13      \\ \hline
\multirow{2}{*}{\begin{tabular}[c]{@{}l@{}}Mixtral-8x22B\\ (Instruct)\end{tabular}} & \multirow{2}{*}{70\%} & \ours (w/ OWL)   & \textbf{30.78} & \textbf{60.20} & 47.95 & 77.86 & 55.41     & 59.56 \\
                                                                                    &                       & OWL                             & 19.64          & 57.74          & 45.48 & 76.60 & 52.47     & 56.42 \\ \hline
\end{tabular}
\caption{Comparison between \ours and the baselines across various models.}
\label{tab:main}
\end{table*}

\begin{table*}[]
\centering
\begin{tabular}{l|c|l|c|c}
\hline
model                                    & sparsity              & method          & cost                      & Avg            \\ \hline
\multirow{3}{*}{Mixtral-8x7B (Instruct)} & 0\%                   & unpruned        &                           & 69.98          \\ \cline{2-5} 
                                         & \multirow{2}{*}{25\%} & Ours            & $O(1)$                    & \textbf{68.05} \\
                                         &                       & \citet{Not2024lu} & $O(\frac{k^n}{\sqrt{n}})$ & 67.45          \\ \hline
\multirow{3}{*}{Mixtral-8x7B}            & 0\%                   & unpruned        &                           & 67.58          \\ \cline{2-5} 
                                         & \multirow{2}{*}{25\%} & Ours            & $O(1)$                    & \textbf{64.34} \\
                                         &                       & \citet{Not2024lu} & $O(\frac{k^n}{\sqrt{n}})$ & 64.22          \\ \hline
\end{tabular}
\caption{Comparing the average performance of 8 tasks of the proposed expert pruning, with other baselines.}
\label{tab:main_structured}
\end{table*}
\begin{table*}[]
\centering
\setlength{\tabcolsep}{1.5pt}
\begin{tabular}{l|c|cccccc|c}
\hline
\textbf{}              & sparsity & \multicolumn{1}{l}{ARC-C} & \multicolumn{1}{l}{BoolQ} & \multicolumn{1}{l}{HellaSwag} & \multicolumn{1}{l}{MMLU} & \multicolumn{1}{l}{RTE} & \multicolumn{1}{l|}{WinoGrande} & \multicolumn{1}{l}{Avg} \\ \hline
Unpruned               & 0\%      & 59.4                      & 84.2                      & 84.0                          & 67.9                     & 70.4                    & 75.6                            & 71.5                    \\ \hline
Ours                   & 25\%     & 55.6                      & 83.1                      & 81.1                          & 63.3                     & 68.6                    & 72.7                            & \textbf{70.7}           \\
SEER-MoE~\cite{SEERMoE2024muzio}        & 25\%     & -                         & -                         & -                             & 56.7                     & -                       & -                               & -                       \\
Expert Drop~\cite{Demystifying2024he}  & 25\%     & 53.2                      & 77.7                      & 80.5                          & 52.2                     & 55.6                    & 76.8                            & 66.0                    \\
Layer Drop~\cite{Demystifying2024he}  & 25\%     & 47.7                      & 85.3                      & 75.2                          & 67.3                     & 69.7                    & 74.6                            & 70.0                    \\ \hline
\end{tabular}
\caption{Comparison of efficient expert pruning methods on Mixtral-8x7B.}
\label{tab:expert_pruning_comparison}
\end{table*}
\begin{table}[]
\centering
\begin{tabular}{l|c|c|c}
\hline
                  & sparsity & cost& GSM8K          \\ \hline
unpruned & 0\% & & 63.46 \\ \hline
Ours              & 25\%     & $O(1)$& \textbf{53.22} \\
\citet{Not2024lu} & 25\%     & $O(\frac{k^n}{\sqrt{n}})$& 48.52          \\ \hline
\end{tabular}
\caption{GSM8K accuracy comparison with baseline on Mixtral-8x7B-Instruct.}
\label{tab:gsm8k_expert_pruning}
\end{table}

\subsection{Experimental Settings}
We use Snowflake Arctic~\cite{SnowflakeLabs2024snowflake} as a representative large MoE, with a total of 480B parameters and 128 experts. To compare our method with previous works~\cite{Not2024lu}, we also experiment with Mixtral~\cite{Mixtral2024jiang}. 

\paragraph{Tasks and Datasets}
In contrast to previous unstructured pruning studies~\cite{Wandasimple2024sun,OWLOutlier2024yin}, we also evaluate the NLG task, GSM8K~\cite{GSM8KTraining2021cobbe}, where maintaining performance proves to be much more challenging (see \autoref{tab:main}; \autoref{sec:gsm8k_harder}).
We further assess performance on four NLU tasks-- ARC-challenge and ARC-easy~\cite{ARCThink2018clark}, HellaSwag~\cite{HellaSwag2019zellers}, and MMLU~\cite{MMLUMeasuring2021hendrycks}.
When comparing with expert pruning methods, following previous work~\cite{Not2024lu}, we also conduct a zero-shot evaluation on BoolQ~\cite{SuperGLUE2019wang}, OpenBookQA~\cite{OpenBookQACan2018mihaylov}, RTE~\cite{GLUE2018wang}, WinoGrande~\cite{WinoGrande2021sakaguchi}.

To provide some data for inference, we employ the C4 dataset~\cite{T5C4Exploring2020raffel}, following the baselines~\cite{OWLOutlier2024yin,Wandasimple2024sun,Not2024lu}. 

\paragraph{Implementation Details}
We explore the values of $(\lambda_1,\lambda_2) \in \{(0,1),(1,0),(1,1)\}$, except for Snowflake Arctic, the largest MoE in our experiments, where we only consider $(\lambda_1,\lambda_2)=(1,0)$, meaning no GPU calls are required for expert pruning.
The sparsity for expert-level structured pruning is set to 20\%, 12.5\%, and 10\% for Snowflake Arctic, Mixtral-8x7B, and Mixtral-8x22B respectively.
We use $\kappa=3$ for selective reconstruction.
More detailed implementation information and hyperparameter decisions are provided in Appendix.

\subsection{Experimental Results}\label{sec:exp_res}
\subsubsection{RQ1: \ours Outperforms Unstructured Pruning}
\autoref{tab:main} describes that our proposed (\ours) significantly outperforms the unstructured pruning methods.
We emphasize that we use the same unstructured pruning approach for all for fair comparison.

For example, with 40\% of sparsity for the Arctic, \ours neatly retains the original GSM8K performance, while unstructured pruning results in a noticeable performance drop.
This is consistent for different unstructured pruning methods, Wanda, as well.
As the sparsity increases, the table shows that \ours can maintain the original performance much better than the baselines-- For 65\% sparsity, \ours's GSM8K performance is nearly 30\%p better than that of unstructured pruning.
With different models, we observe different gaps but a similar trend--
For 65\% of sparsity for Mixtral-8x7B-Instruct, \ours's GSM8K performance is nearly 20 times better than that of unstructured pruning. In the ARC-challenge, the unstructured pruning performance falls below the random-guess accuracy of 25.02~\cite{ARCThink2018clark}, whereas \ours maintains a significantly higher performance, achieving twice the score. Mixtral-8x22B shows a similar trend.

\begin{table}[]
\centering
\begin{tabular}{c|cc}
\hline
sparsity & $O(n)$ & $O(1)$ \\ \hline
25\%      & 63.97  & 64.34  \\
50\%      & 59.90  & 59.58  \\ \hline
\end{tabular}
\caption{Comparing the average performance of 8 tasks of our $O(n)$ and $O(1)$ algorithms for expert pruning on Mixtral-8x7B.}
\label{tab:on_o1}
\end{table}

\begin{figure*}[h]
    \centering
     \subtable[Arctic (128x3.66B)]{%
    \centering
   \includegraphics[width=0.35\textwidth]{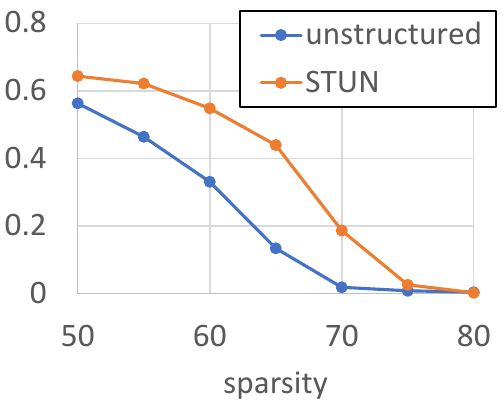}
        }\
    \subtable[Mixtral-8x7B]{%
    \centering
   \includegraphics[width=0.35\textwidth]{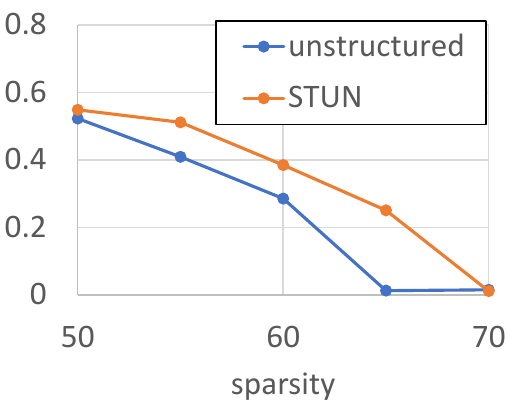}
    }\\
    \subtable[Mixtral-8x22B]{%
    \centering
   \includegraphics[width=0.35\textwidth]{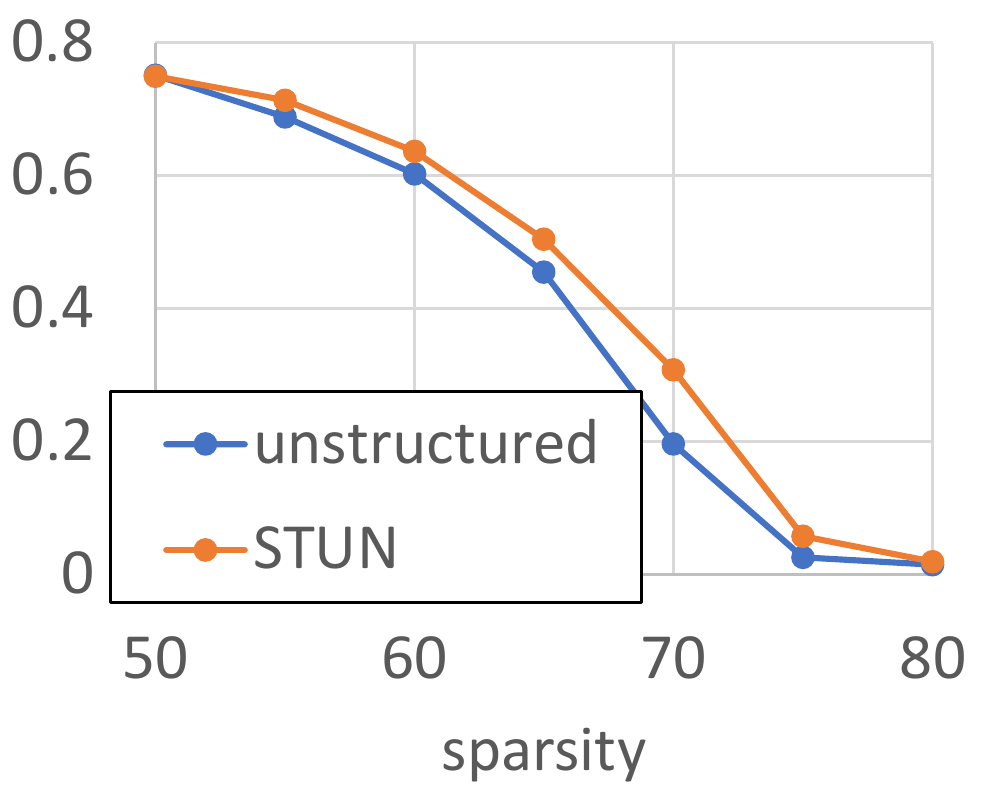}
    }
    \subtable[OLMoE-7B]{%
    \centering
   \includegraphics[width=0.35\textwidth]{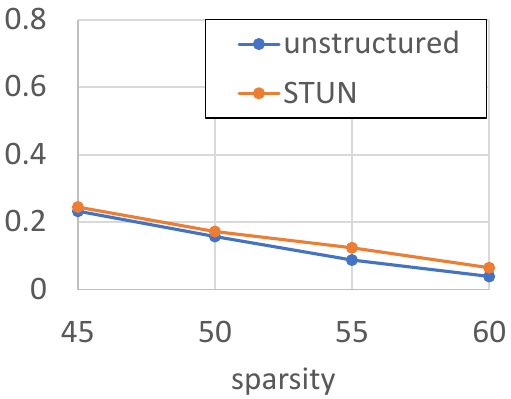}
    }
     \caption{Comparing \ours and unstructured pruning for various MoEs.}
     \label{fig:model_comparison}
\end{figure*}

\begin{figure*}[h]
    \centering
     \subtable[Llama-2 7B]{%
    \centering
   \includegraphics[width=0.48\textwidth]{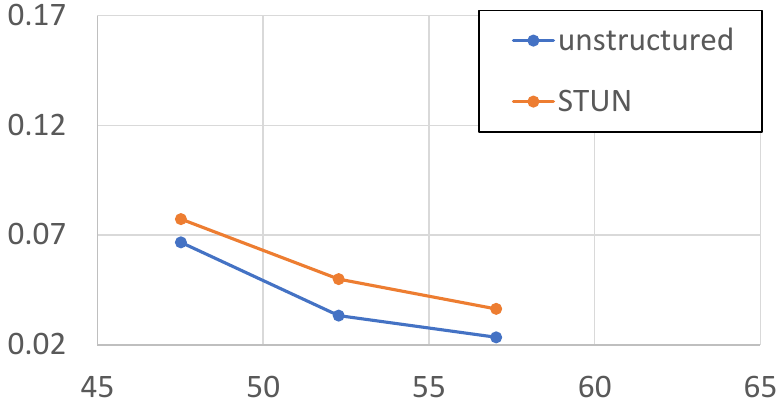}
        }\
    \subtable[Llama-2 13B]{%
    \centering
   \includegraphics[width=0.48\textwidth]{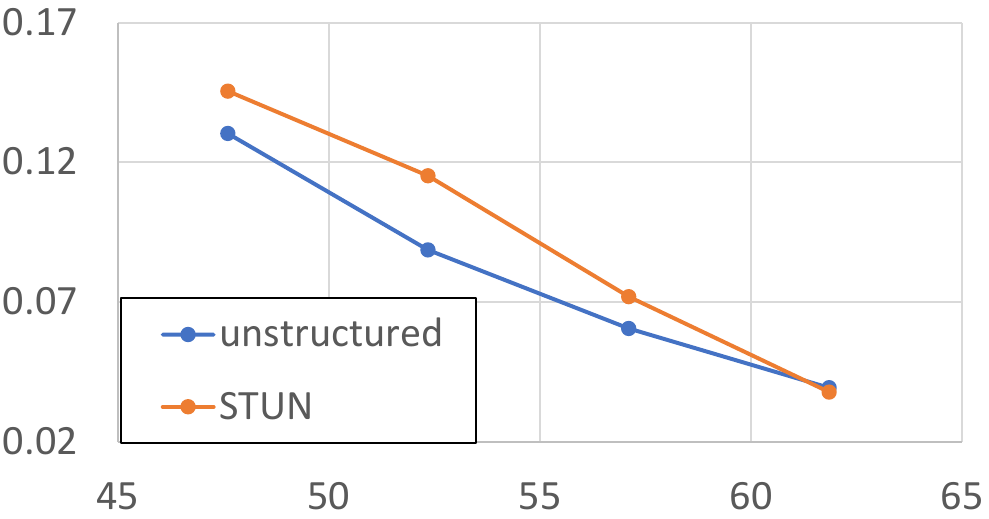}
    }
     \caption{GSM8K 5-shot accuracy comparing \ours and unstructured pruning for various non-MoEs.}
     \label{fig:llama_comparison}
\end{figure*}

\subsubsection{RQ2: Our $O(1)$ Expert Pruning Outperforms Existing Methods}
Tables \ref{tab:main_structured} and \ref{tab:gsm8k_expert_pruning} show that our proposed $O(1)$ expert pruning method is highly effective, outperforming the previous $O(\frac{k^n}{\sqrt{n}})$ solution. 
This is because we derive the latent structure from the pretrained MoE, while the previous work~\cite{Not2024lu} relies solely on the given calibration data.
This validates our design of $O(1)$ in section \ref{sec:ours}. The detailed results are in \autoref{sec:expert_pruning_comparison}.

\autoref{tab:expert_pruning_comparison} shows that our expert pruning outperforms other efficient pruning methods. Compared to SEER-MoE, our pruning clearly outperforms the performance in MMLU, by a substantial margin. Moreover, ours 
achieves a higher average performance compared to the results reported by \citet{Demystifying2024he}.\footnote{Note that SEER-MoE~\cite{SEERMoE2024muzio} only reveals the performance of MMLU, and we used different metrics following \citet{Demystifying2024he}.}

Moreover, \autoref{tab:on_o1} shows that our $O(1)$ expert pruning method achieves similar performance to our $O(n)$ method. This supports our choice to use the $O(1)$ method, which is more efficient.

\begin{table}[]
\centering
\setlength{\tabcolsep}{0.5pt}
\begin{tabular}{l|ccc}
\hline
           & Train & GPU cost & GPUs \# \\ \hline
\citet{Not2024lu} & 0               & infeasible\protect\footnotemark & > 8                  \\
\ours (w/ OWL)      & 0               & 1.12h     & 1                      \\
OWL~\cite{OWLOutlier2024yin}        & 0               & 1.12h     & 1                      \\ 
\ours (w/ Wanda)      & 0               & 0.58h     & 1                      \\
Wanda~\cite{Wandasimple2024sun}      & 0               & 0.58h     & 1                      \\\hline
\end{tabular}
\caption{Cost comparison of diverse pruning methods with Snowflake-Arctic. Train: training cost, GPU \#: number of GPUs required for pruning.}
\label{tab:cost}
\end{table}

\footnotetext{In detail,
23951146041928082866135587776380551750 forward passes
per layer at minimum.}

\subsubsection{RQ3: \ours Adapts Large Number of Small Experts}
\autoref{fig:model_comparison} illustrates the trend of \ours in different MoEs. The performance gap between \ours and unstructured pruning increases as the MoE has more experts with small sizes (from (c) to (a)).
This is because having more experts, rather than having fewer but larger ones, provides greater flexibility to our expert pruning.
Notably, MoEs with a large number of small experts are favored in recent models~\cite{Mixture2024he}.

\subsubsection{RQ4: \ours Outperforms Unstructured Pruning in non-MoEs}
To investigate whether \ours is generalizable to non-MoE as well, we employ a state-of-the-art structured pruning algorithm for non-MoE models, namely, LLM-surgeon~\cite{LLMsurgeonLLM2024vanderouderaa} with 5\% sparsity before performing unstructured pruning, which is OWL in our case. \autoref{fig:llama_comparison} illustrates that such \ours outperforms unstructured pruning.

\section{Cost Analysis}
\autoref{tab:cost} shows the cost comparison between diverse pruning methods. While none of the pruning methods require training cost, \citet{Not2024lu} is infeasible due to its prohibitive number of forward passes in GPUs. Due to the efficiency of proposed expert pruning, \ours is as efficient as the unstructured pruning method it uses, making it a feasible pruning method even for large MoEs, such as Snowflake-Arctic.

\section{Conclusion}

In this paper, we proposed STUN-- an interpolation between structured and unstructured pruning, leveraging both inter- and intra-expert sparsity.%
We provide both theoretical and empirical evidence demonstrating why 
designing expert pruning before unstructured pruning is beneficial.

\section*{Limitation}
Since our method utilizes unstructured pruning in the second stage, we share the same disadvantages with unstructured pruning, that is, on some hardware, the acceleration may not be trivial. 
However, it is shown that some hardware, such as CPU, can successfully accelerate unstructure-pruned networks~\cite{Deepsparseneuralmagic2021neuralmagic}, and ongoing research is actively developing methods to achieve similar speedups on GPUs~\cite{Accelerating2021mishra,exploring2024zhao}. 
With a substantial body of work dedicated to unstructured pruning~\cite{SparseGPT2023frantar,GBLMSize2024das,PrunerZero2024dong,Wandasimple2024sun,OWLOutlier2024yin,Discovering2024li,SSTE2024hu,SPP2024lu}, we believe this approach remains highly relevant and leads to practical performance improvements with ongoing hardware support. 

We already have shown our method works in versatile extreme settings, such as non-MoE models, or high sparsity, etc. We leave more extensive evaluations, such as performance under highly imbalanced or skewed data distributions, as future work.

\bibliography{240615ARR_autogen,custom}

\begin{thebibliography}{67}
\providecommand{\natexlab}[1]{#1}

\bibitem[{Aloise et~al.(2009)Aloise, Deshpande, Hansen, and
  Popat}]{NPhardness2009aloise}
Daniel Aloise, Amit Deshpande, Pierre Hansen, and Preyas Popat. 2009.
\newblock \href {https://doi.org/10.1007/s10994-009-5103-0} {{{NP-hardness}} of
  {{Euclidean}} sum-of-squares clustering}.
\newblock \emph{Machine Learning}, 75(2):245--248.

\bibitem[{Bai et~al.(2023)Bai, Bai, Chu, Cui, Dang, Deng, Fan, Ge, Han, Huang,
  Hui, Ji, Li, Lin, Lin, Liu, Liu, Lu, Lu, Ma, Men, Ren, Ren, Tan, Tan, Tu,
  Wang, Wang, Wang, Wu, Xu, Xu, Yang, Yang, Yang, Yang, Yao, Yu, Yuan, Yuan,
  Zhang, Zhang, Zhang, Zhang, Zhou, Zhou, Zhou, and Zhu}]{Qwen2023bai}
Jinze Bai, Shuai Bai, Yunfei Chu, Zeyu Cui, Kai Dang, Xiaodong Deng, Yang Fan,
  Wenbin Ge, Yu~Han, Fei Huang, Binyuan Hui, Luo Ji, Mei Li, Junyang Lin, Runji
  Lin, Dayiheng Liu, Gao Liu, Chengqiang Lu, Keming Lu, Jianxin Ma, Rui Men,
  Xingzhang Ren, Xuancheng Ren, Chuanqi Tan, Sinan Tan, Jianhong Tu, Peng Wang,
  Shijie Wang, Wei Wang, Shengguang Wu, Benfeng Xu, Jin Xu, An~Yang, Hao Yang,
  Jian Yang, Shusheng Yang, Yang Yao, Bowen Yu, Hongyi Yuan, Zheng Yuan,
  Jianwei Zhang, Xingxuan Zhang, Yichang Zhang, Zhenru Zhang, Chang Zhou,
  Jingren Zhou, Xiaohuan Zhou, and Tianhang Zhu. 2023.
\newblock \href {https://doi.org/10.48550/arXiv.2309.16609} {Qwen {{Technical
  Report}}}.
\newblock \emph{Preprint}, arXiv:2309.16609.

\bibitem[{Behnke and Heafield(2021)}]{Pruning2021behnke}
Maximiliana Behnke and Kenneth Heafield. 2021.
\newblock Pruning {{Neural Machine Translation}} for {{Speed Using Group
  Lasso}}.
\newblock In \emph{Proceedings of the {{Sixth Conference}} on {{Machine
  Translation}}}, pages 1074--1086, Online. Association for Computational
  Linguistics.

\bibitem[{Br{\'e}laz(1979)}]{DSaturNew1979brelaz}
Daniel Br{\'e}laz. 1979.
\newblock \href {https://doi.org/10.1145/359094.359101} {New methods to color
  the vertices of a graph}.
\newblock \emph{Communications of The Acm}, 22(4):251--256.

\bibitem[{Cheng et~al.(2024)Cheng, Zhang, and Shi}]{MINILLM2024cheng}
Hongrong Cheng, Miao Zhang, and Javen~Qinfeng Shi. 2024.
\newblock \href {https://doi.org/10.48550/arXiv.2407.11681} {{{MINI-LLM}}:
  {{Memory-Efficient Structured Pruning}} for {{Large Language Models}}}.
\newblock \emph{Preprint}, arXiv:2407.11681.

\bibitem[{Clark et~al.(2018)Clark, Cowhey, Etzioni, Khot, Sabharwal, Schoenick,
  and Tafjord}]{ARCThink2018clark}
Peter Clark, Isaac Cowhey, Oren Etzioni, Tushar Khot, Ashish Sabharwal, Carissa
  Schoenick, and Oyvind Tafjord. 2018.
\newblock \href {https://doi.org/10.48550/arXiv.1803.05457} {Think you have
  {{Solved Question Answering}}? {{Try ARC}}, the {{AI2 Reasoning Challenge}}}.
\newblock \emph{Preprint}, arXiv:1803.05457.

\bibitem[{Cobbe et~al.(2021)Cobbe, Kosaraju, Bavarian, Chen, Jun, Kaiser,
  Plappert, Tworek, Hilton, Nakano, Hesse, and
  Schulman}]{GSM8KTraining2021cobbe}
Karl Cobbe, Vineet Kosaraju, Mohammad Bavarian, Mark Chen, Heewoo Jun, Lukasz
  Kaiser, Matthias Plappert, Jerry Tworek, Jacob Hilton, Reiichiro Nakano,
  Christopher Hesse, and John Schulman. 2021.
\newblock \href {https://doi.org/10.48550/arXiv.2110.14168} {Training
  {{Verifiers}} to {{Solve Math Word Problems}}}.
\newblock \emph{Preprint}, arXiv:2110.14168.

\bibitem[{Dai et~al.(2024)Dai, Deng, Zhao, Xu, Gao, Chen, Li, Zeng, Yu, Wu,
  Xie, Li, Huang, Luo, Ruan, Sui, and Liang}]{DeepSeekMoE2024dai}
Damai Dai, Chengqi Deng, Chenggang Zhao, R.x. Xu, Huazuo Gao, Deli Chen, Jiashi
  Li, Wangding Zeng, Xingkai Yu, Y.~Wu, Zhenda Xie, Y.k. Li, Panpan Huang, Fuli
  Luo, Chong Ruan, Zhifang Sui, and Wenfeng Liang. 2024.
\newblock {{DeepSeekMoE}}: {{Towards Ultimate Expert Specialization}} in
  {{Mixture-of-Experts Language Models}}.
\newblock In \emph{Proceedings of the 62nd {{Annual Meeting}} of the
  {{Association}} for {{Computational Linguistics}} ({{Volume}} 1: {{Long
  Papers}})}, pages 1280--1297, Bangkok, Thailand. Association for
  Computational Linguistics.

\bibitem[{Darlington(1970)}]{Kurtosis1970darlington}
Richard~B. Darlington. 1970.
\newblock \href {https://doi.org/10.2307/2681925} {Is {{Kurtosis Really}}
  "{{Peakedness}}?"}.
\newblock \emph{The American Statistician}, 24(2):19--22.

\bibitem[{Das et~al.(2024)Das, Sun, Ma, and Shen}]{GBLMSize2024das}
Rocktim~Jyoti Das, Mingjie Sun, Liqun Ma, and Zhiqiang Shen. 2024.
\newblock \href {https://doi.org/10.48550/arXiv.2311.04902} {Beyond {{Size}}:
  {{How Gradients Shape Pruning Decisions}} in {{Large Language Models}}}.
\newblock \emph{Preprint}, arXiv:2311.04902.

\bibitem[{Dasgupta(2008)}]{Dasgupta2008TheHO}
Sanjoy Dasgupta. 2008.
\newblock The hardness of k-means clustering.

\bibitem[{Databricks(2024)}]{databricks2024databricks}
Databricks. 2024.
\newblock Databricks/dbrx.
\newblock Databricks.

\bibitem[{Dery et~al.(2024)Dery, Kolawole, Kagy, Smith, Neubig, and
  Talwalkar}]{Everybody2024dery}
Lucio Dery, Steven Kolawole, Jean-Fran{\c c}ois Kagy, Virginia Smith, Graham
  Neubig, and Ameet Talwalkar. 2024.
\newblock \href {https://doi.org/10.48550/arXiv.2402.05406} {Everybody {{Prune
  Now}}: {{Structured Pruning}} of {{LLMs}} with only {{Forward Passes}}}.
\newblock \emph{Preprint}, arXiv:2402.05406.

\bibitem[{Dettmers et~al.(2023)Dettmers, Pagnoni, Holtzman, and
  Zettlemoyer}]{QLoRA2023dettmers}
Tim Dettmers, Artidoro Pagnoni, Ari Holtzman, and Luke Zettlemoyer. 2023.
\newblock {{QLoRA}}: {{Efficient}} finetuning of quantized {{LLMs}}.
\newblock In \emph{Thirty-Seventh Conference on Neural Information Processing
  Systems}.

\bibitem[{Dong et~al.(2024)Dong, Li, Tang, Liu, Pan, Wang, and
  Chu}]{PrunerZero2024dong}
Peijie Dong, Lujun Li, Zhenheng Tang, Xiang Liu, Xinglin Pan, Qiang Wang, and
  Xiaowen Chu. 2024.
\newblock Pruner-{{Zero}}: {{Evolving Symbolic Pruning Metric From Scratch}}
  for {{Large Language Models}}.
\newblock In \emph{Forty-First {{International Conference}} on {{Machine
  Learning}}}.

\bibitem[{Du et~al.(2022)Du, Huang, Dai, Tong, Lepikhin, Xu, Krikun, Zhou, Yu,
  Firat, Zoph, Fedus, Bosma, Zhou, Wang, Wang, Webster, Pellat, Robinson,
  {Meier-Hellstern}, Duke, Dixon, Zhang, Le, Wu, Chen, and Cui}]{GLaM2022du}
Nan Du, Yanping Huang, Andrew~M Dai, Simon Tong, Dmitry Lepikhin, Yuanzhong Xu,
  Maxim Krikun, Yanqi Zhou, Adams~Wei Yu, Orhan Firat, Barret Zoph, Liam Fedus,
  Maarten~P Bosma, Zongwei Zhou, Tao Wang, Emma Wang, Kellie Webster, Marie
  Pellat, Kevin Robinson, Kathleen {Meier-Hellstern}, Toju Duke, Lucas Dixon,
  Kun Zhang, Quoc Le, Yonghui Wu, Zhifeng Chen, and Claire Cui. 2022.
\newblock {{GLaM}}: {{Efficient}} scaling of language models with
  mixture-of-experts.
\newblock In \emph{Proceedings of the 39th International Conference on Machine
  Learning}, volume 162 of \emph{Proceedings of Machine Learning Research},
  pages 5547--5569. PMLR.

\bibitem[{Fan et~al.(2019)Fan, Grave, and Joulin}]{Reducing2019fan}
Angela Fan, Edouard Grave, and Armand Joulin. 2019.
\newblock Reducing {{Transformer Depth}} on {{Demand}} with {{Structured
  Dropout}}.
\newblock In \emph{International {{Conference}} on {{Learning
  Representations}}}.

\bibitem[{Fedus et~al.(2022)Fedus, Zoph, and Shazeer}]{Switch2022fedus}
William Fedus, Barret Zoph, and Noam Shazeer. 2022.
\newblock Switch transformers: {{Scaling}} to trillion parameter models with
  simple and efficient sparsity.
\newblock \emph{Journal of Machine Learning Research}, 23(120):1--39.

\bibitem[{Frantar and Alistarh(2023)}]{SparseGPT2023frantar}
Elias Frantar and Dan Alistarh. 2023.
\newblock {{SparseGPT}}: {{Massive}} language models can be accurately pruned
  in one-shot.
\newblock In \emph{Proceedings of the 40th International Conference on Machine
  Learning}, volume 202 of \emph{Proceedings of Machine Learning Research},
  pages 10323--10337. PMLR.

\bibitem[{Gao et~al.(2021)Gao, Tow, Biderman, Black, DiPofi, Foster, Golding,
  Hsu, McDonell, Muennighoff, Phang, Reynolds, Tang, Thite, Wang, Wang, and
  Zou}]{eval-harness}
Leo Gao, Jonathan Tow, Stella Biderman, Sid Black, Anthony DiPofi, Charles
  Foster, Laurence Golding, Jeffrey Hsu, Kyle McDonell, Niklas Muennighoff,
  Jason Phang, Laria Reynolds, Eric Tang, Anish Thite, Ben Wang, Kevin Wang,
  and Andy Zou. 2021.
\newblock \href {https://doi.org/10.5281/zenodo.5371628} {A framework for
  few-shot language model evaluation}.
\newblock Zenodo.

\bibitem[{Gao et~al.(2024)Gao, Liu, Zhang, Du, and
  Xia}]{Optimizationbased2024gao}
Yuan Gao, Zujing Liu, Weizhong Zhang, Bo~Du, and Gui-Song Xia. 2024.
\newblock \href {https://doi.org/10.48550/arXiv.2406.10576} {Optimization-based
  {{Structural Pruning}} for {{Large Language Models}} without
  {{Back-Propagation}}}.
\newblock \emph{Preprint}, arXiv:2406.10576.

\bibitem[{Gong et~al.(2022)Gong, Li, and Genzel}]{Adaptive2022gong}
Hongyu Gong, Xian Li, and Dmitriy Genzel. 2022.
\newblock \href {https://arxiv.org/abs/2104.07358} {Adaptive {{Sparse
  Transformer}} for {{Multilingual Translation}}}.
\newblock \emph{arXiv:2104.07358 [cs]}.

\bibitem[{Hassibi and Stork(1992)}]{Second1992hassibi}
Babak Hassibi and David Stork. 1992.
\newblock Second order derivatives for network pruning: {{Optimal Brain
  Surgeon}}.
\newblock In \emph{Advances in {{Neural Information Processing Systems}}},
  volume~5. Morgan-Kaufmann.

\bibitem[{He et~al.(2024)He, Dong, Ding, and Li}]{Demystifying2024he}
Shwai He, Daize Dong, Liang Ding, and Ang Li. 2024.
\newblock \href {https://doi.org/10.48550/arXiv.2406.02500} {Demystifying the
  {{Compression}} of {{Mixture-of-Experts Through}} a {{Unified Framework}}}.
\newblock \emph{Preprint}, arXiv:2406.02500.

\bibitem[{He(2024)}]{Mixture2024he}
Xu~Owen He. 2024.
\newblock \href {https://arxiv.org/abs/2407.04153} {Mixture of {{A Million
  Experts}}}.
\newblock \emph{Preprint}, arXiv:2407.04153.

\bibitem[{Hendrycks et~al.(2021)Hendrycks, Burns, Basart, Zou, Mazeika, Song,
  and Steinhardt}]{MMLUMeasuring2021hendrycks}
Dan Hendrycks, Collin Burns, Steven Basart, Andy Zou, Mantas Mazeika, Dawn
  Song, and Jacob Steinhardt. 2021.
\newblock Measuring massive multitask language understanding.
\newblock In \emph{International Conference on Learning Representations}.

\bibitem[{Hu et~al.(2024)Hu, Zhu, and Chen}]{SSTE2024hu}
Yuezhou Hu, Jun Zhu, and Jianfei Chen. 2024.
\newblock S-{{STE}}: {{Continuous Pruning Function}} for {{Efficient}} 2:4
  {{Sparse Pre-training}}.
\newblock In \emph{The {{Thirty-eighth Annual Conference}} on {{Neural
  Information Processing Systems}}}.

\bibitem[{Jiang et~al.(2023)Jiang, Sablayrolles, Mensch, Bamford, Chaplot,
  de~las Casas, Bressand, Lengyel, Lample, Saulnier, Lavaud, Lachaux, Stock,
  Scao, Lavril, Wang, Lacroix, and Sayed}]{Mistral2023jiang}
Albert~Q. Jiang, Alexandre Sablayrolles, Arthur Mensch, Chris Bamford,
  Devendra~Singh Chaplot, Diego de~las Casas, Florian Bressand, Gianna Lengyel,
  Guillaume Lample, Lucile Saulnier, L{\'e}lio~Renard Lavaud, Marie-Anne
  Lachaux, Pierre Stock, Teven~Le Scao, Thibaut Lavril, Thomas Wang,
  Timoth{\'e}e Lacroix, and William~El Sayed. 2023.
\newblock \href {https://doi.org/10.48550/arXiv.2310.06825} {Mistral {{7B}}}.
\newblock \emph{Preprint}, arXiv:2310.06825.

\bibitem[{Jiang et~al.(2024)Jiang, Sablayrolles, Roux, Mensch, Savary, Bamford,
  Chaplot, de~las Casas, Hanna, Bressand, Lengyel, Bour, Lample, Lavaud,
  Saulnier, Lachaux, Stock, Subramanian, Yang, Antoniak, Scao, Gervet, Lavril,
  Wang, Lacroix, and Sayed}]{Mixtral2024jiang}
Albert~Q. Jiang, Alexandre Sablayrolles, Antoine Roux, Arthur Mensch, Blanche
  Savary, Chris Bamford, Devendra~Singh Chaplot, Diego de~las Casas, Emma~Bou
  Hanna, Florian Bressand, Gianna Lengyel, Guillaume Bour, Guillaume Lample,
  L{\'e}lio~Renard Lavaud, Lucile Saulnier, Marie-Anne Lachaux, Pierre Stock,
  Sandeep Subramanian, Sophia Yang, Szymon Antoniak, Teven~Le Scao,
  Th{\'e}ophile Gervet, Thibaut Lavril, Thomas Wang, Timoth{\'e}e Lacroix, and
  William~El Sayed. 2024.
\newblock \href {https://doi.org/10.48550/arXiv.2401.04088} {Mixtral of
  {{Experts}}}.
\newblock \emph{Preprint}, arXiv:2401.04088.

\bibitem[{Kaddour et~al.(2023)Kaddour, Harris, Mozes, Bradley, Raileanu, and
  McHardy}]{Challenges2023kaddour}
Jean Kaddour, Joshua Harris, Maximilian Mozes, Herbie Bradley, Roberta
  Raileanu, and Robert McHardy. 2023.
\newblock \href {https://doi.org/10.48550/arXiv.2307.10169} {Challenges and
  {{Applications}} of {{Large Language Models}}}.
\newblock \emph{Preprint}, arXiv:2307.10169.

\bibitem[{Kim et~al.(2021)Kim, Awan, Muzio, Salinas, Lu, Hendy, Rajbhandari,
  He, and Awadalla}]{Scalable2021kim}
Young~Jin Kim, Ammar~Ahmad Awan, Alexandre Muzio, Andres Felipe~Cruz Salinas,
  Liyang Lu, Amr Hendy, Samyam Rajbhandari, Yuxiong He, and Hany~Hassan
  Awadalla. 2021.
\newblock \href {https://doi.org/10.48550/arXiv.2109.10465} {Scalable and
  {{Efficient MoE Training}} for {{Multitask Multilingual Models}}}.
\newblock \emph{Preprint}, arXiv:2109.10465.

\bibitem[{Koishekenov et~al.(2023)Koishekenov, Berard, and
  Nikoulina}]{Memoryefficient2023koishekenov}
Yeskendir Koishekenov, Alexandre Berard, and Vassilina Nikoulina. 2023.
\newblock \href {https://doi.org/10.18653/v1/2023.acl-long.198}
  {Memory-efficient {{NLLB-200}}: {{Language-specific Expert Pruning}} of a
  {{Massively Multilingual Machine Translation Model}}}.
\newblock In \emph{Proceedings of the 61st {{Annual Meeting}} of the
  {{Association}} for {{Computational Linguistics}} ({{Volume}} 1: {{Long
  Papers}})}, pages 3567--3585, Toronto, Canada. Association for Computational
  Linguistics.

\bibitem[{Kurtic et~al.(2022)Kurtic, Campos, Nguyen, Frantar, Kurtz, Fineran,
  Goin, and Alistarh}]{oBERTOptimal2022kurtic}
Eldar Kurtic, Daniel Campos, Tuan Nguyen, Elias Frantar, Mark Kurtz, Benjamin
  Fineran, Michael Goin, and Dan Alistarh. 2022.
\newblock \href {https://doi.org/10.18653/v1/2022.emnlp-main.279} {The
  {{Optimal BERT Surgeon}}: {{Scalable}} and {{Accurate Second-Order Pruning}}
  for {{Large Language Models}}}.
\newblock In \emph{Proceedings of the 2022 {{Conference}} on {{Empirical
  Methods}} in {{Natural Language Processing}}}, pages 4163--4181, Abu Dhabi,
  United Arab Emirates. Association for Computational Linguistics.

\bibitem[{Li et~al.(2024)Li, Dong, Tang, Liu, Wang, Luo, Xue, Liu, Chu, and
  Guo}]{Discovering2024li}
Lujun Li, Peijie Dong, Zhenheng Tang, Xiang Liu, Qiang Wang, Wenhan Luo, Wei
  Xue, Qifeng Liu, Xiaowen Chu, and Yike Guo. 2024.
\newblock Discovering {{Sparsity Allocation}} for {{Layer-wise Pruning}} of
  {{Large Language Models}}.
\newblock In \emph{The {{Thirty-eighth Annual Conference}} on {{Neural
  Information Processing Systems}}}.

\bibitem[{Li et~al.(2020)Li, Cooper~Stickland, Tang, and Kong}]{Deep2020li}
Xian Li, Asa Cooper~Stickland, Yuqing Tang, and Xiang Kong. 2020.
\newblock Deep transformers with latent depth.
\newblock In \emph{Advances in Neural Information Processing Systems},
  volume~33, pages 1736--1746. Curran Associates, Inc.

\bibitem[{Liang et~al.(2021)Liang, Zuo, Chen, Jiang, Liu, He, Zhao, and
  Chen}]{Super2021liang}
Chen Liang, Simiao Zuo, Minshuo Chen, Haoming Jiang, Xiaodong Liu, Pengcheng
  He, Tuo Zhao, and Weizhu Chen. 2021.
\newblock \href {https://doi.org/10.18653/v1/2021.acl-long.510} {Super
  {{Tickets}} in {{Pre-Trained Language Models}}: {{From Model Compression}} to
  {{Improving Generalization}}}.
\newblock In \emph{Proceedings of the 59th {{Annual Meeting}} of the
  {{Association}} for {{Computational Linguistics}} and the 11th
  {{International Joint Conference}} on {{Natural Language Processing}}
  ({{Volume}} 1: {{Long Papers}})}, pages 6524--6538, Online. Association for
  Computational Linguistics.

\bibitem[{Lieber et~al.(2024)Lieber, Lenz, Bata, Cohen, Osin, Dalmedigos,
  Safahi, Meirom, Belinkov, {Shalev-Shwartz}, Abend, Alon, Asida, Bergman,
  Glozman, Gokhman, Manevich, Ratner, Rozen, Shwartz, Zusman, and
  Shoham}]{Jamba2024lieber}
Opher Lieber, Barak Lenz, Hofit Bata, Gal Cohen, Jhonathan Osin, Itay
  Dalmedigos, Erez Safahi, Shaked Meirom, Yonatan Belinkov, Shai
  {Shalev-Shwartz}, Omri Abend, Raz Alon, Tomer Asida, Amir Bergman, Roman
  Glozman, Michael Gokhman, Avashalom Manevich, Nir Ratner, Noam Rozen, Erez
  Shwartz, Mor Zusman, and Yoav Shoham. 2024.
\newblock \href {https://doi.org/10.48550/arXiv.2403.19887} {Jamba: {{A Hybrid
  Transformer-Mamba Language Model}}}.
\newblock \emph{Preprint}, arXiv:2403.19887.

\bibitem[{Liu et~al.(2024)Liu, Zhu, Lin, Ning, Blaschko, Yan, Dai, Yang, and
  Wang}]{EEPEfficient2024liu}
Enshu Liu, Junyi Zhu, Zinan Lin, Xuefei Ning, Matthew~B. Blaschko, Shengen Yan,
  Guohao Dai, Huazhong Yang, and Yu~Wang. 2024.
\newblock \href {https://doi.org/10.48550/arXiv.2407.00945} {Efficient {{Expert
  Pruning}} for {{Sparse Mixture-of-Experts Language Models}}: {{Enhancing
  Performance}} and {{Reducing Inference Costs}}}.
\newblock \emph{Preprint}, arXiv:2407.00945.

\bibitem[{Lu et~al.(2024{\natexlab{a}})Lu, Liu, Xu, Zhou, Huang, Zhang, Yan,
  and Li}]{Not2024lu}
Xudong Lu, Qi~Liu, Yuhui Xu, Aojun Zhou, Siyuan Huang, Bo~Zhang, Junchi Yan,
  and Hongsheng Li. 2024{\natexlab{a}}.
\newblock Not {{All Experts}} are {{Equal}}: {{Efficient Expert Pruning}} and
  {{Skipping}} for {{Mixture-of-Experts Large Language Models}}.
\newblock In \emph{Proceedings of the 62nd {{Annual Meeting}} of the
  {{Association}} for {{Computational Linguistics}} ({{Volume}} 1: {{Long
  Papers}})}, pages 6159--6172.

\bibitem[{Lu et~al.(2024{\natexlab{b}})Lu, Zhou, Xu, Zhang, Gao, and
  Li}]{SPP2024lu}
Xudong Lu, Aojun Zhou, Yuhui Xu, Renrui Zhang, Peng Gao, and Hongsheng Li.
  2024{\natexlab{b}}.
\newblock {{SPP}}: {{Sparsity-Preserved Parameter-Efficient Fine-Tuning}} for
  {{Large Language Models}}.
\newblock In \emph{Forty-First {{International Conference}} on {{Machine
  Learning}}}.

\bibitem[{Ma et~al.(2023)Ma, Fang, and Wang}]{LLMPruner2023ma}
Xinyin Ma, Gongfan Fang, and Xinchao Wang. 2023.
\newblock {{LLM-Pruner}}: {{On}} the structural pruning of large language
  models.
\newblock In \emph{Thirty-Seventh Conference on Neural Information Processing
  Systems}.

\bibitem[{{Mason-Williams} and Dahlqvist(2024)}]{What2024mason-williams}
Gabryel {Mason-Williams} and Fredrik Dahlqvist. 2024.
\newblock What makes a good prune? {{Maximal}} unstructured pruning for maximal
  cosine similarity.
\newblock In \emph{The Twelfth International Conference on Learning
  Representations}.

\bibitem[{Megiddo and Supowit(1984)}]{Complexity1984megiddo}
Nimrod Megiddo and Kenneth~J. Supowit. 1984.
\newblock \href {https://doi.org/10.1137/0213014} {On the {{Complexity}} of
  {{Some Common Geometric Location Problems}}}.
\newblock \emph{SIAM Journal on Computing}, 13(1):182--196.

\bibitem[{Mihaylov et~al.(2018)Mihaylov, Clark, Khot, and
  Sabharwal}]{OpenBookQACan2018mihaylov}
Todor Mihaylov, Peter Clark, Tushar Khot, and Ashish Sabharwal. 2018.
\newblock \href {https://doi.org/10.18653/v1/D18-1260} {Can a {{Suit}} of
  {{Armor Conduct Electricity}}? {{A New Dataset}} for {{Open Book Question
  Answering}}}.
\newblock In \emph{Proceedings of the 2018 {{Conference}} on {{Empirical
  Methods}} in {{Natural Language Processing}}}, pages 2381--2391, Brussels,
  Belgium. Association for Computational Linguistics.

\bibitem[{Mishra et~al.(2021)Mishra, Latorre, Pool, Stosic, Stosic, Venkatesh,
  Yu, and Micikevicius}]{Accelerating2021mishra}
Asit Mishra, Jorge~Albericio Latorre, Jeff Pool, Darko Stosic, Dusan Stosic,
  Ganesh Venkatesh, Chong Yu, and Paulius Micikevicius. 2021.
\newblock \href {https://doi.org/10.48550/arXiv.2104.08378} {Accelerating
  {{Sparse Deep Neural Networks}}}.
\newblock \emph{Preprint}, arXiv:2104.08378.

\bibitem[{Muzio et~al.(2024)Muzio, Sun, and He}]{SEERMoE2024muzio}
Alexandre Muzio, Alex Sun, and Churan He. 2024.
\newblock \href {https://doi.org/10.48550/arXiv.2404.05089} {{{SEER-MoE}}:
  {{Sparse Expert Efficiency}} through {{Regularization}} for
  {{Mixture-of-Experts}}}.
\newblock \emph{Preprint}, arXiv:2404.05089.

\bibitem[{NeuralMagic(2021)}]{Deepsparseneuralmagic2021neuralmagic}
NeuralMagic. 2021.
\newblock Neuralmagic/deepsparse: {{Sparsity-aware}} deep learning inference
  runtime for {{CPUs}}.
\newblock https://github.com/neuralmagic/deepsparse.

\bibitem[{OpenAI(2023)}]{GPT42023openai}
OpenAI. 2023.
\newblock \href {https://doi.org/10.48550/arXiv.2303.08774} {{{GPT-4 Technical
  Report}}}.
\newblock \emph{Preprint}, arXiv:2303.08774.

\bibitem[{Raffel et~al.(2020)Raffel, Shazeer, Roberts, Lee, Narang, Matena,
  Zhou, Li, and Liu}]{T5C4Exploring2020raffel}
Colin Raffel, Noam Shazeer, Adam Roberts, Katherine Lee, Sharan Narang, Michael
  Matena, Yanqi Zhou, Wei Li, and Peter~J. Liu. 2020.
\newblock Exploring the limits of transfer learning with a unified text-to-text
  transformer.
\newblock \emph{Journal of Machine Learning Research}, 21(140):1--67.

\bibitem[{Sakaguchi et~al.(2021)Sakaguchi, Bras, Bhagavatula, and
  Choi}]{WinoGrande2021sakaguchi}
Keisuke Sakaguchi, Ronan~Le Bras, Chandra Bhagavatula, and Yejin Choi. 2021.
\newblock \href {https://doi.org/10.1145/3474381} {{{WinoGrande}}: An
  adversarial winograd schema challenge at scale}.
\newblock \emph{Communications of The Acm}, 64(9):99--106.

\bibitem[{Shim et~al.(2021)Shim, Choi, Sung, and Choi}]{Layerwise2021shim}
Kyuhong Shim, Iksoo Choi, Wonyong Sung, and Jungwook Choi. 2021.
\newblock \href {https://doi.org/10.1109/ISOCC53507.2021.9613933} {Layer-wise
  {{Pruning}} of {{Transformer Attention Heads}} for {{Efficient Language
  Modeling}}}.
\newblock In \emph{2021 18th {{International SoC Design Conference}}
  ({{ISOCC}})}, pages 357--358.

\bibitem[{Sneath and Sokal(1973)}]{SingleLinkageNumerical1973sneath}
Peter H.~A. Sneath and Robert~R. Sokal. 1973.
\newblock Numerical taxonomy. {{The}} principles and practice of numerical
  classification.

\bibitem[{Snowflake(2024)}]{SnowflakeLabs2024snowflake}
Snowflake. 2024.
\newblock Snowflake-{{Labs}}/snowflake-arctic.
\newblock Snowflake Labs.

\bibitem[{Strubell et~al.(2019)Strubell, Ganesh, and
  McCallum}]{Energy2019strubell}
Emma Strubell, Ananya Ganesh, and Andrew McCallum. 2019.
\newblock \href {https://doi.org/10.18653/v1/P19-1355} {Energy and {{Policy
  Considerations}} for {{Deep Learning}} in {{NLP}}}.
\newblock In \emph{Proceedings of the 57th {{Annual Meeting}} of the
  {{Association}} for {{Computational Linguistics}}}, pages 3645--3650,
  Florence, Italy. Association for Computational Linguistics.

\bibitem[{Sun et~al.(2024)Sun, Liu, Bair, and Kolter}]{Wandasimple2024sun}
Mingjie Sun, Zhuang Liu, Anna Bair, and J~Zico Kolter. 2024.
\newblock A simple and effective pruning approach for large language models.
\newblock In \emph{The Twelfth International Conference on Learning
  Representations}.

\bibitem[{Touvron et~al.(2023)Touvron, Martin, Stone, Albert, Almahairi,
  Babaei, Bashlykov, Batra, Bhargava, Bhosale, Bikel, Blecher, Ferrer, Chen,
  Cucurull, Esiobu, Fernandes, Fu, Fu, Fuller, Gao, Goswami, Goyal, Hartshorn,
  Hosseini, Hou, Inan, Kardas, Kerkez, Khabsa, Kloumann, Korenev, Koura,
  Lachaux, Lavril, Lee, Liskovich, Lu, Mao, Martinet, Mihaylov, Mishra,
  Molybog, Nie, Poulton, Reizenstein, Rungta, Saladi, Schelten, Silva, Smith,
  Subramanian, Tan, Tang, Taylor, Williams, Kuan, Xu, Yan, Zarov, Zhang, Fan,
  Kambadur, Narang, Rodriguez, Stojnic, Edunov, and
  Scialom}]{LLaMA2Llama2023touvron}
Hugo Touvron, Louis Martin, Kevin Stone, Peter Albert, Amjad Almahairi, Yasmine
  Babaei, Nikolay Bashlykov, Soumya Batra, Prajjwal Bhargava, Shruti Bhosale,
  Dan Bikel, Lukas Blecher, Cristian~Canton Ferrer, Moya Chen, Guillem
  Cucurull, David Esiobu, Jude Fernandes, Jeremy Fu, Wenyin Fu, Brian Fuller,
  Cynthia Gao, Vedanuj Goswami, Naman Goyal, Anthony Hartshorn, Saghar
  Hosseini, Rui Hou, Hakan Inan, Marcin Kardas, Viktor Kerkez, Madian Khabsa,
  Isabel Kloumann, Artem Korenev, Punit~Singh Koura, Marie-Anne Lachaux,
  Thibaut Lavril, Jenya Lee, Diana Liskovich, Yinghai Lu, Yuning Mao, Xavier
  Martinet, Todor Mihaylov, Pushkar Mishra, Igor Molybog, Yixin Nie, Andrew
  Poulton, Jeremy Reizenstein, Rashi Rungta, Kalyan Saladi, Alan Schelten, Ruan
  Silva, Eric~Michael Smith, Ranjan Subramanian, Xiaoqing~Ellen Tan, Binh Tang,
  Ross Taylor, Adina Williams, Jian~Xiang Kuan, Puxin Xu, Zheng Yan, Iliyan
  Zarov, Yuchen Zhang, Angela Fan, Melanie Kambadur, Sharan Narang, Aurelien
  Rodriguez, Robert Stojnic, Sergey Edunov, and Thomas Scialom. 2023.
\newblock \href {https://doi.org/10.48550/arXiv.2307.09288} {Llama 2: {{Open
  Foundation}} and {{Fine-Tuned Chat Models}}}.
\newblock \emph{Preprint}, arXiv:2307.09288.

\bibitem[{{van der Ouderaa} et~al.(2024){van der Ouderaa}, Nagel, Baalen, and
  Blankevoort}]{LLMsurgeonLLM2024vanderouderaa}
Tycho F.~A. {van der Ouderaa}, Markus Nagel, Mart~Van Baalen, and Tijmen
  Blankevoort. 2024.
\newblock The {{LLM}} surgeon.
\newblock In \emph{The Twelfth International Conference on Learning
  Representations}.

\bibitem[{Voita et~al.(2019)Voita, Talbot, Moiseev, Sennrich, and
  Titov}]{Analyzing2019voita}
Elena Voita, David Talbot, Fedor Moiseev, Rico Sennrich, and Ivan Titov. 2019.
\newblock \href {https://doi.org/10.18653/v1/P19-1580} {Analyzing {{Multi-Head
  Self-Attention}}: {{Specialized Heads Do}} the {{Heavy Lifting}}, the {{Rest
  Can Be Pruned}}}.
\newblock In \emph{Proceedings of the 57th {{Annual Meeting}} of the
  {{Association}} for {{Computational Linguistics}}}, pages 5797--5808,
  Florence, Italy. Association for Computational Linguistics.

\bibitem[{Wang et~al.(2019)Wang, Pruksachatkun, Nangia, Singh, Michael, Hill,
  Levy, and Bowman}]{SuperGLUE2019wang}
Alex Wang, Yada Pruksachatkun, Nikita Nangia, Amanpreet Singh, Julian Michael,
  Felix Hill, Omer Levy, and Samuel~R. Bowman. 2019.
\newblock {{SuperGLUE}}: A stickier benchmark for general-purpose language
  understanding systems.
\newblock In \emph{Proceedings of the 33rd {{International Conference}} on
  {{Neural Information Processing Systems}}}, 294, pages 3266--3280. Curran
  Associates Inc., Red Hook, NY, USA.

\bibitem[{Wang et~al.(2018)Wang, Singh, Michael, Hill, Levy, and
  Bowman}]{GLUE2018wang}
Alex Wang, Amanpreet Singh, Julian Michael, Felix Hill, Omer Levy, and Samuel
  Bowman. 2018.
\newblock \href {https://doi.org/10.18653/v1/W18-5446} {{{GLUE}}: {{A
  Multi-Task Benchmark}} and {{Analysis Platform}} for {{Natural Language
  Understanding}}}.
\newblock In \emph{Proceedings of the 2018 {{EMNLP Workshop BlackboxNLP}}:
  {{Analyzing}} and {{Interpreting Neural Networks}} for {{NLP}}}, pages
  353--355, Brussels, Belgium. Association for Computational Linguistics.

\bibitem[{Yin et~al.(2024)Yin, Wu, Zhang, Hsieh, Wang, Jia, Li, Jaiswal,
  Pechenizkiy, Liang, Bendersky, Wang, and Liu}]{OWLOutlier2024yin}
Lu~Yin, You Wu, Zhenyu Zhang, Cheng-Yu Hsieh, Yaqing Wang, Yiling Jia, Gen Li,
  Ajay~Kumar Jaiswal, Mykola Pechenizkiy, Yi~Liang, Michael Bendersky,
  Zhangyang Wang, and Shiwei Liu. 2024.
\newblock Outlier {{Weighed Layerwise Sparsity}} ({{OWL}}): {{A Missing Secret
  Sauce}} for {{Pruning LLMs}} to {{High Sparsity}}.
\newblock In \emph{Forty-First {{International Conference}} on {{Machine
  Learning}}}.

\bibitem[{Zellers et~al.(2019)Zellers, Holtzman, Bisk, Farhadi, and
  Choi}]{HellaSwag2019zellers}
Rowan Zellers, Ari Holtzman, Yonatan Bisk, Ali Farhadi, and Yejin Choi. 2019.
\newblock \href {https://doi.org/10.18653/v1/P19-1472} {{{HellaSwag}}: {{Can}}
  a {{Machine Really Finish Your Sentence}}?}
\newblock In \emph{Proceedings of the 57th {{Annual Meeting}} of the
  {{Association}} for {{Computational Linguistics}}}, pages 4791--4800,
  Florence, Italy. Association for Computational Linguistics.

\bibitem[{Zeng et~al.(2023)Zeng, Garay, Zhou, Chong, Hua, Wu, Pan, Zhou, Voigt,
  and Yang}]{GreenPLM2023zeng}
Qingcheng Zeng, Lucas Garay, Peilin Zhou, Dading Chong, Yining Hua, Jiageng Wu,
  Yikang Pan, Han Zhou, Rob Voigt, and Jie Yang. 2023.
\newblock \href {https://doi.org/10.24963/ijcai.2023/698} {{{GreenPLM}}:
  {{Cross-Lingual Transfer}} of {{Monolingual Pre-Trained Language Models}} at
  {{Almost No Cost}}}.
\newblock In \emph{Proceedings of the {{Thirty-Second International Joint
  Conference}} on {{Artificial Intelligence}}}, pages 6290--6298, Macau, SAR
  China. International Joint Conferences on Artificial Intelligence
  Organization.

\bibitem[{Zhang et~al.(2024{\natexlab{a}})Zhang, XiaolongShi, Sun, and
  Sun}]{Structured2024zhang}
Honghe Zhang, XiaolongShi XiaolongShi, Jingwei Sun, and Guangzhong Sun.
  2024{\natexlab{a}}.
\newblock \href {https://doi.org/10.18653/v1/2024.findings-naacl.1} {Structured
  {{Pruning}} for {{Large Language Models Using Coupled Components
  Elimination}} and {{Minor Fine-tuning}}}.
\newblock In \emph{Findings of the {{Association}} for {{Computational
  Linguistics}}: {{NAACL}} 2024}, pages 1--12, Mexico City, Mexico. Association
  for Computational Linguistics.

\bibitem[{Zhang et~al.(2024{\natexlab{b}})Zhang, Liu, Cheng, Xu, and
  Gao}]{Diversifying2024zhang}
Zeliang Zhang, Xiaodong Liu, Hao Cheng, Chenliang Xu, and Jianfeng Gao.
  2024{\natexlab{b}}.
\newblock Diversifying the {{Expert Knowledge}} for {{Task-Agnostic Pruning}}
  in {{Sparse Mixture-of-Experts}}.

\bibitem[{Zhang et~al.(2021)Zhang, Qi, Liu, Liu, and Sun}]{Know2021zhang}
Zhengyan Zhang, Fanchao Qi, Zhiyuan Liu, Qun Liu, and Maosong Sun. 2021.
\newblock \href {https://doi.org/10.1016/j.aiopen.2021.05.003} {Know what you
  don't need: {{Single-Shot Meta-Pruning}} for attention heads}.
\newblock \emph{AI Open}, 2:36--42.

\bibitem[{Zhao et~al.(2024)Zhao, Yuan, Bao, Su, Gao, Sun, Liang, Jing, and
  Chen}]{exploring2024zhao}
Kang Zhao, Tao Yuan, Han Bao, Zhenfeng Su, Chang Gao, Zhaofeng Sun, Zichen
  Liang, Liping Jing, and Jianfei Chen. 2024.
\newblock \href {https://doi.org/10.48550/arXiv.2410.16135} {Beyond 2:4:
  Exploring {{V}}:{{N}}:{{M}} sparsity for efficient transformer inference on
  {{GPUs}}}.
\newblock \emph{Preprint}, arXiv:2410.16135.

\end{thebibliography}

\clearpage
\appendix
\section{Derivation From $O(\frac{k^n}{\sqrt{n}})$ to $O(1)$}
\subsection{Towards $O(n)$: Probabilistic Interpretation}\label{sec:O_n}
As a stepping stone towards $O(1)$, we propose to
rephrase the goal of finding $\theta_S$ to minimize $\mathcal{E}_S$ (Eq. \ref{eq:reconstruction_loss}) as:
\begin{equation}
\textrm{argmax}_S P(X_1=s_1, \cdots , X_{|S|}=s_{|S|})\label{eq:prob_combination}
\end{equation}
where $s_i$s are the experts included in the expert set $S$, and 
$P(X_1=s_1, \cdots , X_{|S|}=s_{|S|})$ is the joint probability of pruning $S$.
One intuitive way to design $P$ so that Eq. \ref{eq:prob_combination} yields the same $S$ to minimize Eq. \ref{eq:reconstruction_loss} is as follows:

\begin{equation}\label{eq:joint_prob_construction}
P(X_1=s_1, \cdots , X_{|S|}=s_{|S|}) = \frac{1}{Z} \cdot \frac{1}{\mathcal{E}_S}
\end{equation}
where $Z$ is the normalization factor. Each joint probability needs $O(1)$ GPU calls, since $\mathcal{E}_S$ needs the output of $M(x;\theta-\theta_S)$.

Section \ref{sec:combinatorial_reconstruction} corresponds to
 enumerating joint probability from all combinations, requiring $n\choose{|S|}$ different values, which is compute-intensive.
 When chain rule is applied,  Eq. \ref{eq:prob_combination} can be reformulated as follows:
\begin{multline}
\textrm{argmax}_S \prod_k P(X_k=s_k|X_1=s_1,\\
\cdots,X_{k-1}=s_{k-1})\label{eq:prob_conditional}
\end{multline}

Our contribution is greedy optimization without compromise for Eq. \ref{eq:prob_conditional}%
-- We decompose the multiplication of Eq.~\ref{eq:prob_conditional} at each step $k$, and obtain the distribution $P(X_k|s_1,\cdots,s_{k-1})$, to select $X_k$ that maximizes the probability.
For simplicity, we will omit $X_k$s from this point on.

As our goal is finding the argmax of the probabilities as in Eq. \ref{eq:prob_conditional}, estimating the rank between them
is sufficient, rather than evaluating exact values. Such rank estimation can benefit from the latent structure among experts, specifically, a cluster of similar experts in MoE, enabling $P(X_k|s_1,\cdots,s_{k-1})$ calculation without chain-rule multiplications in Eq. \ref{eq:prob_conditional}.

Assume we know oracle clusters, $c(E_i)$,
where $c$ is the mapping from an expert to a set of similarly behaving experts identified from the latent clusters.
When we have knowledge of similar experts, 
for example, $c(E_i)=c(E_j)=\{E_i,E_j\}$ 
indicating $E_i$ and $E_j$ are highly similar,
we will decide not to  prune $E_i$ 
if 
$E_j$ is already pruned.
That is, if $c(E_i) \subseteq S_k$ then $P(X_k=E_i|S_{k-1})$ should be lowered by some value $p$, to guide the model against pruning.
Moreover, $P(E_i|S_{k-1})$ should be larger, or rank higher, otherwise. 

To generalize, we will cluster similar experts. Once the cluster of similar experts is finalized, we assign the value $P(E_i|S_{k-1})$,
as follows:

\begin{equation}\label{eq:O_n_prob}
P(E_i|S_{k-1}) = \begin{cases}
    P(E_i) - p & c(E_i) \subseteq S_{k} \\
    P(E_i) & \mathrm{otherwise}
    \end{cases}
\end{equation}
We set $p$ as a constant for simplicity.
This enables the calculation of all $P(E_i|S_{k-1})$ in Eq. \ref{eq:prob_conditional} from $P(E_i)$s, which needs only $n$ forwards in total.

\paragraph*{Clustering the Similar Experts} 
Our remaining task is to obtain
cluster information $c$: %
One signal is pairwise behavioral similarity $b_{i,j}$,
from the pretrained weights $W$ at a minimal cost.
Suppose two rows $W_{i} \approx W_{j}$ are similar; then $r_i(x) \approx r_j(x)$, meaning $E_i,E_j$ tend to be selected by similar inputs, implying similar expertise.
Thus, %
the behavioral similarity $b_{i,j}$ between two experts $E_i,E_j$ can be obtained as follows:

\begin{equation}
b_{i,j} = - \|W_{i} - W_{j}\|_F\label{eq:sim_weight}
\end{equation}

Next, we generalize pairwise similarity %
into clusters of experts, such that experts in each cluster $C_l$
are highly similar to its representative $\mu_l$.
Formally, the objective of clustering is to minimize
the sum of squared errors between $\mu_{l}$ and experts $E_i$ in the cluster: %
\begin{equation}
\sum_{i \in C_l}\sum_l \|W_{i} - \mu_{l}\|^2
\end{equation}
which is an NP-hard problem~\cite{Complexity1984megiddo,Dasgupta2008TheHO,NPhardness2009aloise}.

Practically, we found that the agglomerative clustering algorithm~\cite{SingleLinkageNumerical1973sneath} performs well.\footnote{We tried other clustering algorithms in the Appendix.}
Specifically, clusters are initialized
as individual experts and then iteratively merged,
with a termination condition 
that prevents the experts within each cluster from being too dissimilar. This condition is tuned based on the desired sparsity.

Lastly, if we allow inference on some data, we can improve Eq. \ref{eq:sim_weight} with coactivation statistics $a_{i,j}$, which measure the frequency with which $E_i,E_j$ are selected simultaneously.\footnote{We normalize $a_{i,j}$ by dividing it with the total coactivations in one layer.}
However, these coactivation statistics depend on the given data, whose distribution may differ from the test data.
Therefore, we combine the two as follows:
\begin{equation}
b_{i,j} = - \lambda_1 \|W_{i} - W_{j}\|_F + \lambda_2 a_{i,j}
\end{equation}
We recap the algorithm in the Appendix (Alg \ref{alg:cluster}).

\subsection{Towards $O(1)$: Taylor Approximation and Selective Reconstruction}\label{sec:O_1}
\paragraph*{1st-order Taylor Approximation}
While previous section immensely reduces the cost to obtain the probability distribution to $O(n)$ by requiring only $P(E_i)$s, we can 
further reduce the number of forward passes-- We aim to remove the GPU calls for $P(E_i)$, which is needed as in Eq. \ref{eq:joint_prob_construction}.

The key idea is approximating $E_i$'s reconstruction loss value $\mathcal{E}_i=\|M(x;\theta)-M(x;\theta-\theta_i)\|_F$, and assigning $P(E_i)$ as some high value $L$ if the reconstruction loss $\mathcal{E}_i$ is lowest. This neatly estimates the rank between $P(E_i)$s.

Though $\mathcal{E}_i$ can be approximated via conventional 2nd-order reconstruction methods~\cite{Second1992hassibi,SparseGPT2023frantar}, the size of the hessian matrix increases quadratically with the number of experts, which often yields out-of-memory errors.

To address this, we propose using a 1st order Taylor approximation. To rank the reconstruction loss values, we consider approximating the reconstruction loss when replacing the output from $\theta_i$ with some specific expert $\theta_C$ in $C=c(E_i)$ as follows:
\begin{equation}
\mathcal{E}_i = \|E'(\theta_i) \cdot (\theta_i - \theta_C )\|^2
\end{equation}
As the convention of 2nd order Taylor approximation~\cite{Second1992hassibi,SparseGPT2023frantar}, we assume the parameters are near a local minimum. 
Thus, with a small constant $\gamma$, $\|E'(\theta_i)\| < \gamma$, leading to:
\begin{equation}
\sum_i \mathcal{E}_i < \sum_i \gamma \|\theta_i - \theta_C \|^2\label{eq:upper_bound}
\end{equation}
whose upper bound in the right-hand side can be minimized when $\theta_C = \overline{\theta_i}$, where $\bar{~}$ denotes the average.

Therefore, the expert closest to $\bar{\theta_i}$ within each cluster has the highest priority to be retained. We assign $\mathcal{E}_i$ a large number $L>p$ if $E_i$ is the closest to $\bar{\theta_i}$ from the corresponding cluster $c(E_i)$, and set it to zero otherwise.
The same greedy algorithm is applied to optimize Eq. \ref{eq:prob_conditional}.

\paragraph*{Selective Reconstruction of Experts}
While letting $\theta_C$ as the expert closest to $\bar{\theta_i}$ successfully minimizes $\sum_i \mathcal{E}_i$, sometimes we can minimize them further, by replacing the weight of the closest expert $\theta_C$ to $\bar{\theta_i}$. However, blindly doing so
is suboptimal, as there is another kind of error to consider.
The decision boundaries of the next layer are accustomed to the output of
$\{E(x;\theta_i)\}_{i=1}^{|C|}$, but changing the output as $E(x;\theta_C)=E(x;\bar{\theta_i})$ could result in a distribution that the model is unfamiliar with. This potential error, which we denote as $\mathcal{E}_d$, would be minimized if $\theta_C \in \{\theta_i\}_{i=1}^{|C|}$.

To balance these two types of errors, we selectively decide whether to reconstruct. We observe that $\sum_i \mathcal{E}_i$ increases if the total number of clusters in a layer decreases, as this would introduce more $\|E'(\theta_i) \cdot (\theta_i - \theta_C)\|^2$ terms. Therefore, if the total number of clusters is below a threshold $\kappa$, we use $\theta_C = \overline{\theta_i}$ to minimize $\sum_i \mathcal{E}_i$. 
Otherwise, we set $\theta_C$ as the expert within the cluster $\{\theta_i\}_{i=1}^{|C|}$ closest to the $\overline {\theta_i}$, to minimize $\mathcal{E}_d$.
The router weight reconstruction is done similarly, following its corresponding expert.

\section{Implementation Details}
We probe $(\lambda_1,\lambda_2) \in \{(0,1),(1,0),(1,1)\}$, except for the Snowflake Arctic, which is the biggest MoE we deal with, where we only consider $(\lambda_1,\lambda_2)=(1,0)$, which means no GPU calls is needed for expert pruning.
To get coactivation values $a_{i,j}$, we utilize 1000 samples from the C4 dataset, each of which has 2048 sequence length.
We evaluate on \textsc{lm-evaluation-harness}~\cite{eval-harness}
We use 4bit quantization~\cite{QLoRA2023dettmers} for experiments with Mixtral-8x22B and Arctic, due to their model size. 
We use 20\%, 12.5\%, and 10\% for Arctic, Mixtral-8x7B, and Mixtral-8x22B respectively, as the expert sparsity for \ours. These are the maximum values among 10, 12.5, 20, 25, and 35\%, with minimum performance loss.
We use $\kappa=3$ for selective reconstruction.
For Wanda and OWL, we use 128 C4 samples following the original papers~\cite{OWLOutlier2024yin,Wandasimple2024sun}, while we use 4096 for sequence length. For OWL, we use the default setting, $M=5, \lambda=0.08$.

All experiments are conducted on H100 80GB GPUs, with a maximum of 4. Each evaluation is done within 4 hours, and each unstructured pruning requires less than 2 hours on one GPU. Evaluation is done only once, since we introduce no randomness in our experiment.

\section{Other Clustering Algorithms}
We also considered DSatur~\cite{DSaturNew1979brelaz} as a clustering algorithm for Eq. 12, converting into clique-partitioning in a graph where  edge $e_{i,j}$ connected if two experts are similar enough as follows, 
\begin{equation}
e_{i,j} = \begin{cases}
    1 & b_{i,j} >= t_{DSatur}\\
    \infty & \mathrm{otherwise}
    \end{cases}
\end{equation}
where $t_{DSatur}$ is some threshold to control the sparsity of MoE.

\begin{table}[]
\centering
\begin{tabular}{cc|c}
\hline
Cluster & Reconstruct & LM-eval Avg    \\ \hline
Ours  & Ours        & \textbf{59.58} \\
DSatur                & Ours        & 58.59          \\
Ours  & Always                    & 57.60          \\
Ours  & Never                     & 59.22          \\ \hline
\end{tabular}
\caption{Ablation experiments for the first component of \ours, the proposed expert pruning.}
\label{tab:ablation}
\end{table}

\section{Ablation Studies}
To validate our design of expert pruning in sections \ref{sec:O_n} and \ref{sec:O_1}, we evaluate alternative approaches to expert-prune Mixtral-8x7B at 50\% sparsity.
\autoref{tab:ablation} confirms that our design choices are valid. Our agglomerative clustering algorithm outperforms the DSatur algorithm, an alternative clustering algorithm we discuss in the Appendix.
Additionally, selective reconstruction proves superior to always or never reconstructing, as shown in the last two rows.
Detailed per-task performance of ablation studies are described in Tables \ref{tab:clustering}, \ref{tab:kappa_merging}. Detailed hyperparameter ablation studies are described in Tables \ref{tab:hyp_search}, \ref{tab:struct_sparsity_search}. Results with applying expert-pruning only are provided in \autoref{tab:structured_only}.
\begin{table*}[]
\centering
\setlength{\tabcolsep}{2pt}
\begin{tabular}{l|cccccccc|c}
\hline
                 & ARC-c & ARC-e & BoolQ & HellaSwag & MMLU  & OBQA  & RTE   & WinoGrande & Avg            \\ \hline
DSatur & 45.99 & 75.38 & 80.76 & 54.62     & 45.06 & 29.00 & 66.79 & 71.11      & 58.59          \\
Ours            & 45.73 & 75.13 & 83.46 & 54.55     & 53.29 & 31.20 & 62.45 & 70.80      & \textbf{59.58} \\ \hline
\end{tabular}
\caption{Our agglomerative clustering algorithm is better than the alternative.}
\label{tab:clustering}
\end{table*}
\begin{table*}[]
\centering
\setlength{\tabcolsep}{2pt}
\begin{tabular}{l|cccccccc|c}
\hline
    & ARC-c & ARC-e & BoolQ & HellaSwag & MMLU  & OBQA  & RTE   & WinoGrande & Avg            \\ \hline
Always reconstruct    ($\kappa=8$)  & 42.92 & 74.20 & 82.42 & 54.66     & 50.09 & 28.40 & 55.96 & 72.14      & 57.60          \\
No reconstruct    ($\kappa=0$)  & 45.22 & 75.21 & 82.45 & 54.53     & 52.31 & 30.00 & 62.82 & 71.19      & 59.22          \\
Ours       ($\kappa=3$) & 45.73 & 75.13 & 83.46 & 54.55     & 53.29 & 31.20 & 62.45 & 70.80      & \textbf{59.58} \\ \hline
\end{tabular}
\caption{Selective reconstruction outperforms the baselines.}
\label{tab:kappa_merging}
\end{table*}

\begin{table}[]
\centering
\begin{tabular}{l|cc|c}
\hline
                               & $\lambda_1$ & $\lambda_2$ & GSM8K          \\ \hline
\multirow{3}{*}{Mixtral-8x7B}  & 1           & 0           & \textbf{63.38} \\
                               & 0           & 1           & 58.53          \\
                               & 1           & 1           & 60.42          \\ \hline
\multirow{3}{*}{Mixtral-8x22B} & 1           & 0           & 81.50          \\
                               & 0           & 1           & \textbf{81.58} \\
                               & 1           & 1           & 80.52          \\ \hline
\end{tabular}
\caption{Comparison of different $\lambda_1$, $\lambda_2$ configurations. For the Arctic, we only consider the cheapest, ($\lambda_1$, $\lambda_2$) = (1,0).}
\label{tab:hyp_search}
\end{table}
\begin{table}[]
\centering
\begin{tabular}{l|c|c}
\hline
                              & sparsity & GSM8K \\ \hline
\multirow{3}{*}{Arctic}       & 10\%     & 70.74 \\
                              & 20\%     & 69.90 \\
                              & 35\%     & 60.05 \\ \hline
\multirow{2}{*}{Mixtral-8x7B} & 12.50\%  & 63.38 \\
                              & 25\%     & 53.22 \\ \hline
\multirow{3}{*}{Mixtral-22B}  & 10\%     & 81.58 \\
                              & 12.50\%  & 79.91 \\
                              & 25\%     & 74.07 \\ \hline
\end{tabular}
\caption{Comparison of different sparsity of our proposed expert-level pruning. ($\lambda_1$, $\lambda_2$) is probed among {(1,0),(0,1),(1,1)}, except for 10\% and 20\% of Arctic.}
\label{tab:struct_sparsity_search}
\end{table}
\begin{table*}[]
\centering
\setlength{\tabcolsep}{1pt}
\resizebox{\textwidth}{!}{%
\begin{tabular}{l|c|l|c|cccccccc|c}
\hline
model                                                                              & sparsity             & method            & cost                      & ARC-c & ARC-e & BoolQ & HellaSwag & MMLU  & OBQA  & RTE   & WinoGrande & Avg            \\ \hline
\multirow{3}{*}{\begin{tabular}[c]{@{}l@{}}Mixtral-8x7B\\ (Instruct)\end{tabular}} & 0\%                   & unpruned          &                           & 62.20 & 87.04 & 88.50 & 67.59     & 68.87 & 36.60 & 72.20 & 76.87      & 69.98          \\ \cline{2-13} 
                                                                                   & \multirow{2}{*}{25\%} & Ours              & $O(1)$                    & 59.30 & 85.44 & 88.13 & 64.42     & 64.52 & 35.40 & 71.84 & 75.37      & \textbf{68.05} \\
                                                                                   &                      & \citet{Not2024lu} & $O(\frac{k^n}{\sqrt{n}})$ & 58.19 & 84.89 & 87.34 & 65.24     & 62.47 & 35.60 & 70.04 & 75.85      & 67.45          \\ \hline
\multirow{3}{*}{Mixtral-8x7B}                                                      & 0\%                   & unpruned          &                           & 57.17 & 84.01 & 85.35 & 64.88     & 67.88 & 35.00 & 70.40 & 75.93      & 67.58          \\ \cline{2-13} 
                                                                                   & \multirow{2}{*}{25\%} & Ours              & $O(1)$                    & 52.73 & 81.82 & 83.09 & 60.84     & 63.34 & 31.60 & 68.59 & 72.69      & \textbf{64.34} \\
                                                                                   &                      & \citet{Not2024lu} & $O(\frac{k^n}{\sqrt{n}})$   & 51.62 & 81.94 & 83.64 & 61.60     & 58.72 & 33.00 & 67.87 & 75.37      & 64.22          \\ \hline
\end{tabular}%
}
\caption{Comparing the first component of \ours, the proposed expert pruning, with other baselines.}
\label{tab:vs_luetal_more}
\end{table*}
\begin{table*}[]
\centering
\begin{tabular}{l|c|c|ccccc}
\hline
              & sparsity & GSM8K & Avg (\textrightarrow)   & ARC-c & ARC-e & HellaSwag & MMLU  \\ \hline
Arctic        & 20\%     & 69.90  & 67.95 & 57.00    & 83.71 & 65.70      & 65.39 \\
Mixtral-8x7B  & 12.50\%  & 63.38 & 70.65 & 61.18 & 86.66 & 66.80      & 67.97 \\
Mixtral-8x22B & 10\%     & 81.58 & 71.78 & 61.18 & 85.82 & 66.62     & 73.52 \\ \hline
\end{tabular}
\caption{Performance when applying only our expert pruning.}
\label{tab:structured_only}
\end{table*}

\section{Detailed Results for RQ2}\label{sec:expert_pruning_comparison}
\autoref{tab:vs_luetal_more} describes the per-task performance of RQ2.

\section{Why GSM8K is Harder}\label{sec:gsm8k_harder}
The nature of the GSM8K task, which is a generation task, accounts for this discrepancy. A random generation baseline on GSM8K would achieve 0\% accuracy, making it far more challenging to maintain performance. In contrast, ARC, HellaSwag, and MMLU are multiple-choice tasks where random baselines can achieve reasonable accuracy by comparing the perplexity of different completion options. As a result, maintaining performance on GSM8K is considerably harder, a challenge often overlooked in previous works. This explains the more pronounced difference in performance between STUN and the baselines on GSM8K.

\end{document}